\documentclass[11pt,twocolumn]{article}
\usepackage[margin=0.75in]{geometry}
\usepackage{setspace}
\usepackage{titlesec}
\titlespacing*{\section}{0pt}{1.5ex plus 1ex minus .2ex}{1ex plus .2ex}
\titlespacing*{\subsection}{0pt}{1.25ex plus 1ex minus .2ex}{0.75ex plus .2ex}
\titlespacing*{\subsubsection}{0pt}{1ex plus 1ex minus .2ex}{0.5ex plus .2ex}
\usepackage{algorithmic}
\usepackage{algorithm}
\usepackage{array}
\usepackage[caption=false,font=normalsize,labelfont={sf,scriptsize},textfont=sf]{subfig}
\usepackage{textcomp}
\usepackage{amsmath,amsfonts}
\usepackage{url}
\usepackage{verbatim}
\usepackage{graphicx}
\usepackage{booktabs}
\usepackage{float}
\usepackage{makecell}
\usepackage{tcolorbox}
\usepackage[hidelinks,bookmarksopen,bookmarksnumbered]{hyperref}
\usepackage[authoryear,round]{natbib}

\begin{document}

\title{Beyond the ``Truth": Investigating Election Rumors on Truth Social During the 2024 Election}

\author{Etienne Casanova$^{1,*}$ and R. Michael Alvarez$^1$\\
$^1$California Institute of Technology, Pasadena, CA 91125 USA\\
$^*$ecasanov@caltech.edu}

\date{}

\maketitle

\begin{abstract}
Large language models (LLMs) offer unprecedented opportunities for analyzing social phenomena at scale. This paper demonstrates the value of LLMs in psychological measurement by (1) compiling the first large-scale dataset of election rumors on a niche alt-tech platform, (2) developing a multistage Rumor Detection Agent that leverages LLMs for high-precision content classification, and (3) quantifying the psychological dynamics of rumor propagation, specifically the ``illusory truth effect'' in a naturalistic setting. The Rumor Detection Agent combines (i) a synthetic data-augmented, fine-tuned RoBERTa classifier, (ii) precision keyword filtering, and (iii) a two-pass LLM verification pipeline using GPT-4o mini. The findings reveal that sharing probability rises steadily with each additional exposure, providing large-scale empirical evidence for dose-response belief reinforcement in ideologically homogeneous networks. Simulation results further demonstrate rapid contagion effects: nearly one quarter of users become ``infected'' within just four propagation iterations. Taken together, these results illustrate how LLMs can transform psychological science by enabling the rigorous measurement of belief dynamics and misinformation spread in massive, real-world datasets.
\end{abstract}

\textbf{Keywords:} Rumor detection, rumor propagation, Truth Social, election integrity, machine learning, network analysis, misinformation

\section{Introduction}
Social media platforms increasingly dominate political discourse, amplifying both information and rumors at unprecedented rates. Accurately identifying specific rumor narratives within millions of unstructured posts presents a significant methodological challenge. By leveraging Large Language Models (LLMs), we address these measurement hurdles and provide a scalable framework for investigating election rumors on Truth Social.

The 2020 U.S. presidential election notably demonstrated the prevalence of false claims and their significant impact on public perceptions of electoral integrity \citep{Enders2021}. In response to Twitter suspending Donald Trump, Truth Social emerged in 2021 as a self-described ``free-speech'' alternative, predominantly attracting conservative users and followers of President Donald Trump. By late 2024, the platform hosted a significant user base, creating a distinctively homogeneous ideological environment optimal for studying rumor dynamics.

Truth Social is operated by Trump Media \& Technology Group (TMTG) and was launched on 21 February 2022. Functionally, the platform mirrors X's (formerly Twitter) timeline model: users publish short posts called \emph{Truths} and reposts others' Truths to their own accounts via \emph{ReTruths}. Throughout this paper we adopt the platform's nomenclature: referring to original posts as Truths and to reshares as ReTruths.

Previous research underscores the importance of platform-specific characteristics in rumor dissemination, yet few studies have quantitatively explored the psychological dynamics of rumor spread in an ideologically homogeneous network like Truth Social. This paper addresses this gap by providing the first fine-grained analysis of election rumors on Truth Social and analyzing their propagation through the network. We build upon existing quantitative methodologies, particularly network-based rumor propagation models, to analyze the extent and mechanisms of rumor spread during the 2024 U.S. presidential election.

By leveraging a unique dataset of nearly 15 million posts from approximately 200,000 Truth Social users, we systematically quantify rumors with a rigorous, multi-stage Rumor Detection Agent. This agent exemplifies the value of LLMs in psychological measurement, combining the efficiency of traditional classifiers with the reasoning capabilities of GPT-4o mini to validate claims at scale. Our analysis specifically focuses on the network structures facilitating rumor propagation, providing insights into the roles that influential actors play within ideologically homogeneous platforms. Understanding these dynamics is essential not only for understanding contemporary political communication, but also for informing measures that might mitigate the social impacts of rumors.

\subsection{Past Work}
This section reviews relevant literature in four key areas that inform our research, providing specific insights into rumor dynamics and computational detection approaches. We organize the review into four parts: (i) how rumors spread on mainstream vs. alt-tech platforms, (ii) computational detection frameworks utilizing LLMs, (iii) Truth Social-specific studies, and (iv) propagation models and psychological drivers.

\subsubsection{Rumors on Other Platforms}
A substantial amount of research has examined how political rumors spread on mainstream social media. Research has shown that false news spreads more rapidly and broadly than true news on platforms like Twitter \citep{Vosoughi2018}, while fake election stories during the 2016 U.S. presidential race circulated extensively among a concentrated group of highly active users \citep{Grinberg2019}. During the 2016 U.S. presidential race, for example, ``fake news'' circulated extensively on Twitter and similarly influenced discourse on Facebook \citep{Allcott2017}. In the 2020 election cycle, major platforms tightened content moderation, but this prompted many disaffected users to migrate to alternative sites such as Parler and Gab that promised fewer restrictions. Studies find that these alt-tech platforms, including Truth Social, became echo chambers with minimal oversight, often serving as hotbeds for rumors and conspiracy theories \citep{Zannettou2018,Aliapoulios2021}. \citet{DelVicario2016PNAS} showed that Facebook misinformation thrives within segmented echo chambers, where confirmation bias accelerates false narratives. \citet{Shao2018} demonstrated that social bots disproportionately amplify low-credibility content, especially in early diffusion stages. \citet{HortaRibeiro2023} found that Parler's January 2021 deplatforming did not reduce user activity; instead, users dispersed to other fringe platforms, showing that alt-tech misinformation networks are highly resilient to single-site takedowns. \citet{Cinelli2021} quantified that echo-chamber effects on Gab exceed those on mainstream sites, reinforcing how homogeneous alt communities exacerbate misinformation propagation. This broader literature highlights that rumors spread readily in online political discourse, particularly in partisan communities, setting the stage for our focused analysis of Truth Social's unique characteristics.

\subsubsection{Computational Approaches to Rumor Detection}
The computational detection of misinformation and rumors has evolved considerably over the past decade, driven by advances in natural language processing and machine learning. Early approaches relied heavily on feature engineering, extracting linguistic and metadata features from social media posts to train traditional classifiers, although accuracy varied with training data and context. Recently, LLMs have emerged as strong tools for detecting rumors. In this paper, we utilize a combination of pre-trained LLMs and smaller fine-tuned transformers for rumor detection.

A major branch of rumor detection represents a post together with its reply or repost cascade as a tree and learns a rumor label from both the content and the propagation structure. For example, \citet{Ma2020TreeRumorDetection} introduced bottom-up and top-down tree-structured recursive neural networks with attention to capture stance signals and temporal order in these cascades, and showed that conversation structure substantially improves classification accuracy on Twitter-style data.

Recent work has begun to use large language models directly for rumor or fact-verification tasks. \citet{Zeng2025LLMRumorDetection} find that while LLMs can reason about rumors from social context, their performance degrades when the context is long and highly structured, and they require task-specific orchestration to stay reliable. \citet{Zhang2023LLMFactVerification} address this by decomposing a news claim into verifiable subclaims through a hierarchical step-by-step prompting method, achieving performance comparable to supervised models in few-shot settings. These studies show that LLMs can act as flexible veracity checkers when the claim is well-formed and context is manageable.

To cope with breaking events where labeled data is scarce, \citet{Zhang2024MultiAgentDebate} propose a stance-separated multi-agent debate framework in which LLM agents first separate supporting and opposing comments and then debate to reach a final verdict. In parallel, \citet{Ghosh2023EarlyMisinformation} design an early misinformation detector that leverages very initial propagation paths plus linguistic cues to issue a prediction before a cascade fully forms. Together with the recent survey on LLMs for fake-news detection by \citet{Kuntur2025SurveyLLMFakeNews}, this line of work highlights that modern rumor detection increasingly mixes conversational signals, LLM reasoning, and lightweight propagation features.

Complementing automated detection, agencies and fact-checking organizations have developed rumor validation frameworks to systematically debunk false narratives. Notably, the U.S. Cybersecurity and Infrastructure Security Agency launched a ``Rumor vs. Reality'' resource to fact-check prevalent election myths in real time \citep{CISA2024}. This effort cataloged common false claims (e.g., Dirty Voter Rolls or Ballot Mail-In Fraud) alongside evidence-based corrections, providing a knowledge base. Some recent studies integrate multi-stage pipelines combining keyword filtering, machine-learning classification, and automated verification techniques to improve detection precision \citep{Kochkina2018}. Research has shown that correcting misinformation and rumor remains challenging, as false beliefs often persist even after exposure to accurate information \citep{Ecker2022}, though inoculation and pre-bunking have shown promise as pre-emptive techniques to neutralize rumors before they spread widely \citep{Roozenbeek2022}.

Recent work by \citet{Raza2025FakeNews} demonstrates that fine-tuned BERT-style models (e.g., BERT, RoBERTa) often exceed GPT-3.5 in supervised fake-news classification accuracy, while GPT-based annotators can generate high-quality training data that boosts downstream performance. Similarly, \citet{Hu2024BadAdvisor} found that GPT-3.5 serves better as an ``advisor'' (providing rationales) than as a standalone classifier, and that distilling these LLM rationales into smaller transformers yields state-of-the-art detection systems.

Taken together, these strands of work show that state-of-the-art rumor detection increasingly relies on conversational structure, stance diversity, LLM-based reasoning, and access to rich platform metadata. In our case, several of these assumptions do not fully hold on Truth Social. Conversations on the platform only occasionally form the deep, mixed-stance reply trees that tree-structured models are designed for, which limits the benefit of propagation-based approaches. The platform's largely homogeneous ideological environment also reduces the opposing-view signals that agent-based debate methods expect when separating supporting and refuting comments. Finally, our goal is to assign rumor labels to posts at platform scale, so running multi-step claim decomposition or long-context LLM orchestration for every post would add dramatic cost for little gain. For these reasons, we adopt a lightweight, multi-stage pipeline that is calibrated on Truth Social language, aligned with the CISA taxonomy, and optimized for high-precision labeling over a large corpus. Additionally, we aim to make this framework transferable to other platforms and settings that require large-scale rumor labeling against a fixed set of rumor categories.

\subsubsection{Truth Social: Data and Studies}
Because Truth Social is a relatively new platform, research specific to it has only just begun to emerge. \citet{Gerard2023} provide one of the first large-scale datasets of Truth Social content, compiling over 823,000 posts from approximately 454,000 users and offering basic analysis of the platform's content and network structure. Their dataset helped illuminate the platform's user community and interaction patterns, addressing the early lack of data availability. More recently, \citet{Shah2024} released a comprehensive 2024 election discourse dataset with 1.5 million Truth Social posts related to the presidential campaign. In their study, Truth Social is characterized as an ``unfiltered'' space with minimal content moderation, which has facilitated vibrant political discussion but also the spread of conspiratorial narratives. This line of work provides valuable descriptive insights that inform our understanding of Truth Social's ecosystem, for instance, documenting the prevalence of election-related keywords, user engagement trends, and the prominence of extremist content.

In addition to data contributions, researchers have developed infrastructure to support Truth Social analyses. In particular, \citet{Shah2024} developed the Truthbrush tool, an open-source API client that enables programmatic retrieval of posts and user data from Truth Social's backend. Taken together, these efforts establish an early foundation of knowledge about Truth Social and its role in the online rumor landscape, though there is still much to be explored. Our study is the first to move beyond descriptive statistics and model the platform's rumor diffusion in depth.

\subsubsection{Rumor Propagation}
There have also been efforts to demonstrate the speed and scale of election falsehoods on mainstream social media. For example, \citet{Vosoughi2018} showed that false news spreads significantly faster and farther than true news on Twitter, while \citet{Grinberg2019} found that fake election stories on Twitter were shared by a concentrated minority of highly active users. Similarly, \citet{Allcott2017} quantified the reach of fake news during the 2016 election (e.g., estimating its exposure relative to real news). 

Past work by \citet{Berinksy2023} underscores how political elites seed and legitimize rumors, showing with survey and experimental evidence that conspiratorial predispositions and partisan loyalty drive rumor acceptance - and that among ordinary Republicans co-partisan endorsements boost receptivity. What remains unexamined, however, is how these elite-initiated election rumors propagate ``in the wild'', especially on highly partisan, niche platforms. Our study fills that gap by tracing how election rumors originate, spread, and gain traction on Truth Social, and by quantifying the catalytic role partisan elites play in those cascades.

Finally, emerging research is modeling misinformation spread in novel ways that underscore its broader impact. \citet{DeVerna2025} developed an agent-based simulation framework to quantify how online falsehoods can amplify real-world phenomena. In their model, misinformed social media users are incorporated into an epidemic disease transmission simulation through a ``susceptible-misinformed-infected-recovered (SMIR)'' model, extending the classic SIR epidemiological model. This data-informed model demonstrates, for example, that a higher prevalence of misinformation (such as health or election rumors) in an information network can lead to worse outcomes in a parallel physical network (more disease spread). Such work, while not specific to Truth Social, provides sophisticated quantitative tools to study the psychological dynamics of rumor propagation.

\subsection{Hypothesis}
Building on evidence that repeated exposure to political messaging enhances engagement and diffusion in partisan networks \citep{Cohen2019, Berinksy2023}, we propose the following hypotheses for Truth Social.

A user's probability of sharing a given election-related rumor increases with each additional prior exposure to that rumor. Formally, let $G = (V, E)$ be a directed, weighted graph where each node $u \in V$ is a Truth Social user and each edge $(v \to u) \in E$ represents an impression of rumor from user $v$ to user $u$, weighted by $v$'s interaction frequency. We define $P_{u,r}^{\text{share}}(k)$ as the probability that user $u$ will share rumor $r$ upon receiving exactly their $k$-th exposure to that rumor. We hypothesize that for any rumor $r$ and exposure count $k$:

\begin{equation}
P_{u,r}^{\text{share}}(k+1) > P_{u,r}^{\text{share}}(k), \quad \forall k \ge 0.
\label{eq:hypothesis_exposure}
\end{equation}

This hypothesis formalizes the exposure-propagation relationship by asserting that each additional rumor impression raises a user's probability of sharing that rumor. Intuitively, repeated exposures reinforce message salience and lower users' skepticism, making them more prone to pass the claim along. This cumulative effect mirrors classic social-infection models, in which successive contacts amplify diffusion momentum. We will test this hypothesis in Section~\ref{sec:testing_exposure_relationship} by estimating exposure-propagation curves across users with varying exposure histories.

This dose-response relationship is supported by cognitive research showing that repeated exposure increases perceived accuracy of false statements, a phenomenon known as the “illusory truth effect” \citep{Pennycook2021PsychFakeNews}. Swire-Thompson and Lazer further note that echo chambers increase this effect by reinforcing confirmation biases, making repeated rumors particularly potent in closed communities \citep{SwireThompson2020HealthMisinformation}.

\section{Data Collection}
\subsection{Dataset Overview}
Data scraping began on 24 September 2024 and ended on 31 December 2024. The dataset contains posts ranging from 14 May 2023 to 31 December 2024, with the majority of content concentrated from September 2024 onward. The dataset includes nearly 15 million posts from 200,000 users. After labeling, the dataset contains just under 100,000 rumor posts. Our focus on the period around Election Day, spans from 24 September 2024 to 1 December 2024.

\subsection{Data Scraping Methods}
\begin{figure*}
\centering
\includegraphics[width=\textwidth]{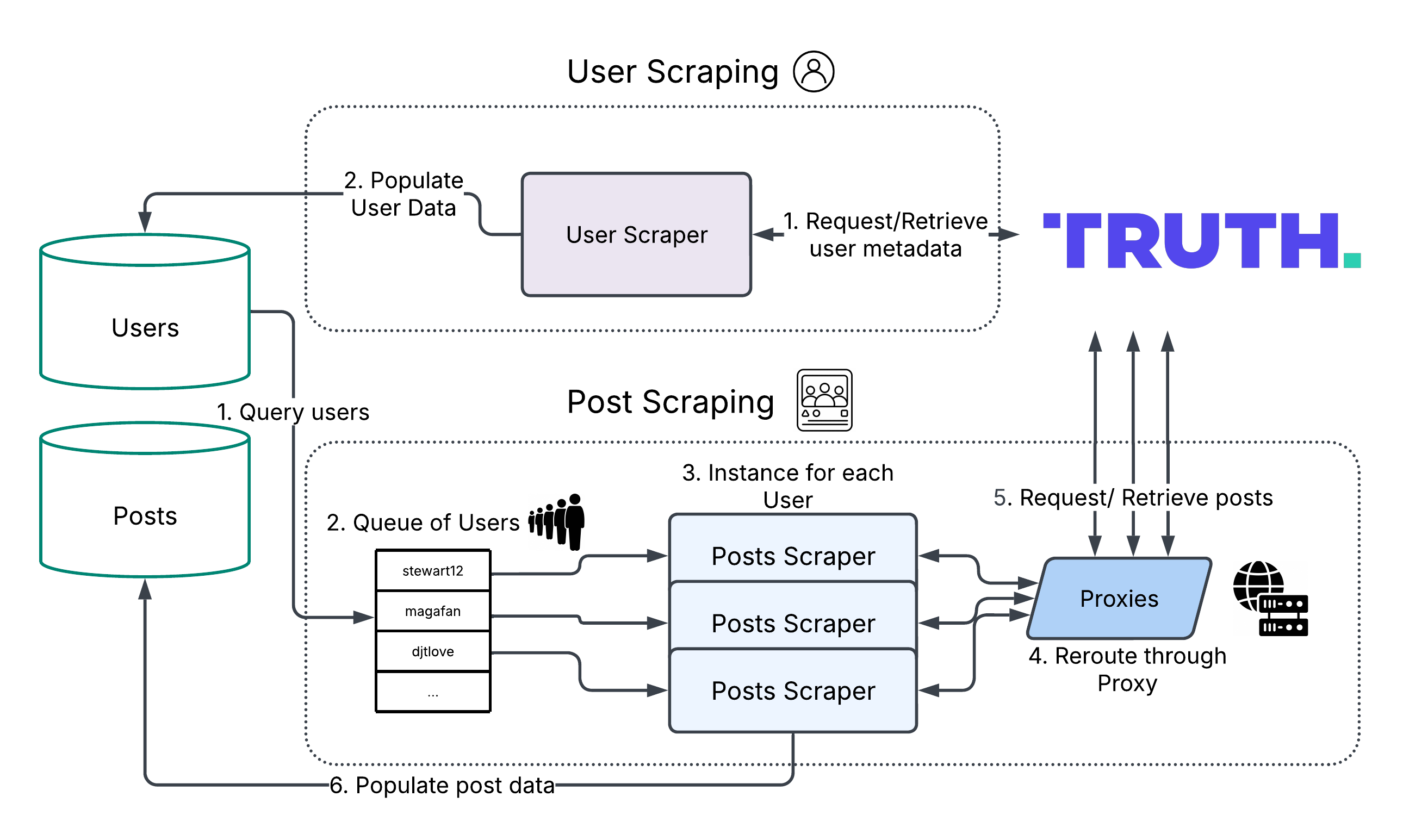}
\caption{Webscraping user metadata and posts from Truth Social's website. The database schema is detailed in Figure S1.}
\label{fig:webscraper}
\end{figure*}

We developed a comprehensive web scraping system to collect Truth Social data, shown in Figure~\ref{fig:webscraper}. Our approach used a distributed network of proxy servers and authenticated accounts to systematically gather user profiles and posts while respecting rate limits. This approach for collecting data from new and alt social media platforms was reviewed by the Institutional Review Board (IRB) at the California Institute of Technology and determined to be exempt (IR24-1473). 

For all requests to Truth Social, we used Truthbrush \citep{Shah2024}. Proxies were used to distribute requests, ensuring efficiency and avoiding rate limiting. The data scraping was done in the following order (each step relied on the previous): 1. Scraping of users 2. Scraping of posts 3. Geolocation of users. Because conversation on the platform centers on Donald Trump, we initiated a breadth-first crawl from his @realDonaldTrump account to obtain a representative sample of the platform's user distribution. We scraped 190,445 total users, from which 83.87\% of active users posted at least once about political discourse. Additionally, we scraped 14,871,193 posts from these users. The algorithms for user and post scraping are detailed in the Supplementary Materials (Algorithms S1 and S2, respectively).

Recent data-collection studies of alternative social networks demonstrate that network-level statistics generalize with datasets of this scale. For Truth Social, Gerard et al. publish a snapshot of 454,000 users with 823,000 posts \citep{Gerard2023}. Similarly, \citet{Lima2018} analyze 171,000 Gab users with 12.9 million posts, and \citet{Zannettou2018} examine 336,000 Gab users with 22 million posts.

Seeding the user scraping at @realDonaldTrump may raise the concern that we've introduced a sampling bias favoring Trump as the network's central node. In practice, we employ two design choices to mitigate this bias: (1) We limit every account to at most 20 followers and 20 following, so the scraper never plunges deeply into Trump's millions of followers. Instead, it quickly branches out to newly discovered users, preventing any one node's local neighborhood from overwhelming the sample. (2) Breadth-first search algorithms naturally visit high degree nodes first - so high degree accounts are sampled early in the search. This allows for fair reach comparisons between top nodes, and dilutes any advantage from being the seed. Together, these safeguards ensure that Trump's measured influence derives from the genuine propagation of his rumors on Truth Social, rather than from artifacts of our scraping methodology.

\subsection{User Geolocation}
To enable geographic analysis of rumor propagation, we developed a multi-stage geolocation pipeline focusing on state-level assignment, as most accounts lack sufficient information for county-level precision. Our approach prioritizes the most reliable geolocation sources, utilizing an LLM intermediary (OpenAI's GPT-4o mini with 0 temperature for replicable and deterministic results) when needed to parse locations from text, handling typos, state nicknames, and city names to produce clean state labels.

The pipeline assigns relative confidence scores based on data source reliability: metadata (1.0), username (0.9), phrase detection (0.6), frequency analysis (0.2), and friend-based inference (0.1). These scores, determined through manual sampling and accuracy estimation, guide the prioritization of geolocation methods. A detailed description of the geolocation process is provided in the Supplementary Materials (Figure S2).

Of our 190,445 users, we successfully geolocated 63,951 (33.58\%). The majority were located through metadata (52,918 users, 27.79\%), followed by friend-based inference (8,451 users, 4.44\%), username analysis (1,772 users, 0.93\%), frequency analysis (441 users, 0.23\%), and phrase detection (369 users, 0.19\%). While 126,494 users (66.42\%) could not be reliably geolocated, only 37,114 (19.48\%) of these un-locatable users had any posts in their accounts, suggesting many were inactive or minimal-content profiles.

\section{Rumor Detection}
\subsection{Defining Election Rumors}
Determining the precise definition of election rumors is crucial for accurately detecting them. For the case of this paper, we have decided to tightly bound rumors, meaning that a post is only considered rumor if it directly contradicts one or more factual claims. In particular, we focus on the list of pre-election and Election Day rumors and realities defined by the Cybersecurity and Infrastructure Security Agency (CISA). A comprehensive list of these rumor categories is provided in the Supplementary Materials (Table S1).

If a post simply jokes about or references the topic of a rumor without explicitly contradicting the facts, we do not label it as rumor. This stricter boundary leaves many rumor-hinting posts unlabeled, but it keeps the dataset cleaner by excluding ambiguous cases.

\subsection{Rumor Detection Agent}
\begin{figure*}
\centering
\includegraphics[width=\textwidth]{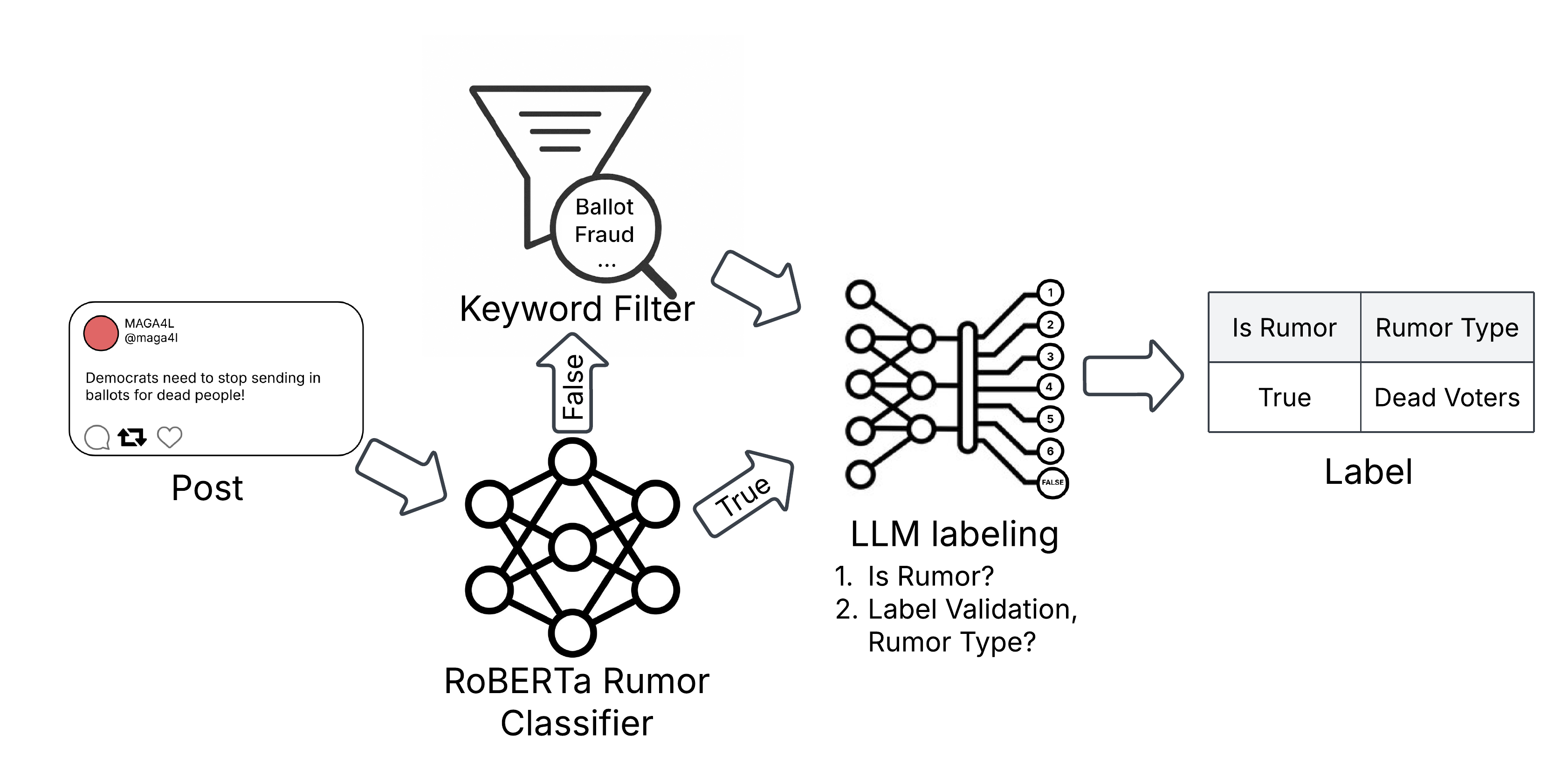}
\caption{Rumor Detection Agent used to classify Truth Social posts. The system uses a fine-tuned RoBERTa model and keyword filtering to identify posts with high probabilities of being a rumor. The posts are then sent to an LLM (GPT-4o mini) where few shot prompting is used first to classify the post as rumor. A second prompt then ensures that the post corresponds to at least one specific rumor.}
\label{fig:misinfo_pipeline}
\end{figure*}

To identify and label rumors at scale, we developed a novel multi-stage Rumor Detection Agent tailored to the unique content and terminology of Truth Social. The agent balances accuracy and efficiency by combining supervised machine learning with rule-based filtering and large language model (LLM) validation, as illustrated in Figure~\ref{fig:misinfo_pipeline}. The complete implementation is available online.\footnote{\url{https://github.com/etmaca5/RumorDetectionAgent}}

Our multi-stage setup is designed for efficiency: we let a fine-tuned RoBERTa model and keyword filters handle most posts, and send only the remaining, potential rumor posts to the LLM. This works particularly well in a platform ecosystem, where the fraction of potential rumor posts is so low. Empirically, only approximately 6.8\% of posts pass through the RoBERTa and keyword filtering stages to reach the LLM verification step, reducing LLM API calls by over 93\% compared to a naive approach that would process every post with an LLM. This substantial cost reduction makes the system practical for processing millions of Truth Social posts while preserving high precision, which is necessary when processing large-scale datasets rather than a small set of individual claims.

The first stage employs a RoBERTa-based classifier. A pre-trained RoBERTa model was fine-tuned on a dataset of mixed synthetic and real posts. For synthetic data generation, we took hand-labeled samples of varied Truth Social posts, both rumors and non-rumors, and generated a dataset of 5000 examples of each type using few-shot prompting with GPT-4o mini. Importantly, this synthetic data is used only for training the RoBERTa filter; the final LLM verification stage operates exclusively on real Truth Social posts, ensuring no circularity between data generation and validation. This resulted in a model that did not generalize well, but was excellent at classifying the typical language of rumor posts, serving well as a coarse-grained filter that scores each post for the likelihood of being a rumor. The RoBERTa classifier effectively filters out the majority of non-rumor posts. In parallel to this classifier, we also employ keyword filtering with hundreds of keywords related to each of the CISA rumor categories to ensure we do not miss any rumor posts that the RoBERTa classifier might overlook.

In the final stage, we leverage a small, few-shot prompted LLM (GPT-4o mini) to first decide the label for posts from the previous steps by asking: ``Is this post a rumor, given the list of known rumors?'' Then we validate that determination with a different query, asking the LLM for deeper certainty that the post is indeed a rumor by asking: ``Are you certain the post corresponds to the description of at least one of the rumors?'' The aim of having two-step verification is to significantly decrease the number of false positives, as we do not want to classify a post as a false claim if it is merely generalizing or being vague about a rumor. This design echoes findings by \citet{Zellers2019NeuralFakeNews}, who showed that large generative models can both produce and detect sophisticated fake news, achieving over 90\% discrimination accuracy when pitted against human-written text. It also aligns with the two-stage paradigm of \citet{Hu2024BadAdvisor} where LLM explanations inform a downstream transformer, improving precision over either model alone.

\subsection{Validation of Rumor Detection Accuracy}
To validate the performance of our Rumor Detection Agent methodology, we conducted human validation using a stratified random sample. A total of 500 posts were extracted using simple random sampling, with 250 labeled rumor posts and 250 labeled non-rumor posts from our post-labeled dataset. These were randomly ordered and presented to a trained human annotator who was provided with detailed validation instructions.

Our validation results demonstrate strong overall performance of our Rumor Detection Agent (detailed metrics are provided in Table S2). The agent achieved 94.60\% accuracy in correctly identifying rumor posts, with perfect recall (100\%) ensuring that no election rumor was missed. This suggests our multi-stage approach efficiently captures rumors without false negatives, although it does produce a modest 10.80\% false positive rate. It is crucial to understand that this 10.80\% false positive rate applies to the balanced validation context, not to the full dataset where rumors constitute only 0.66\% of posts. When applied to the actual dataset prevalence, the positive predictive value (PPV) of our classifier is approximately 89.2\%, meaning that about 89\% of posts flagged as rumors by our agent are indeed rumors. For rumor type classification, the system achieved 91.48\% accuracy, with particularly strong performance for the most prevalent rumor types. These results parallel the human-in-the-loop pipeline of \citet{Mendes2023HITL} for early COVID-19 misinformation, which combined automated claim detection with expert review to achieve high precision in identifying emergent falsehoods. They highlight the benefit of integrating algorithmic filters with overseers, which is mirrored in our LLM-verification step.

Qualitative analysis of the false positives reveals that the majority stem from ambiguous cases that lie at the boundary of our strict rumor definition. Many of these posts make general assertions about election system vulnerability without directly asserting a specific false claim that contradicts CISA's documented realities. For instance, a post claiming that ``Raffensperger knows his election system is wide open... and that results can be manipulated at will'' was flagged by our agent but not labeled as a rumor by human annotators, as it makes speculative claims about system vulnerability rather than stating an explicit falsehood. Such posts operate in a gray zone where the distinction between rumor and non-rumor depends on interpretation. This pattern suggests that our agent's false positives are not random errors but rather systematic detections of content that could reasonably be argued to constitute rumors under a more permissive classification scheme.

\section{Results}
\subsection{Basic Statistics}
Table~\ref{tab:dataset_summary} summarizes the overall structure of our dataset, which includes over 12 million posts and 2 million reposts from nearly 200,000 users. While original rumor posts account for just 0.66\% of total posts, they represent a disproportionately high share of repost activity: the ratio of rumor reTruths to all reTruths is 9.31, compared to an overall reTruth-to-Truth ratio of 0.227. This suggests that rumors are substantially more likely to be amplified than typical content on the platform. We also note that of the 63,744 users with at least one post, 17,239 of those users posted about an election rumor, meaning that 27\% of active users during the election timeframe either Truthed or ReTruthed an election rumor. By comparison, during the 2016 U.S. presidential campaign only about 8.5\% of Facebook users shared links from untrustworthy news sites \citep{Guess2019}.

\begin{table*}
\small\sf\centering
\caption{Summary of Dataset Statistics}
\label{tab:dataset_summary}
\begin{tabular}{ll}
\toprule
Metric & Count \\
\midrule
Users Scraped & 190,445 \\
Original Posts (Truths) & 12,120,620 \\
Reposts (ReTruths) & 2,750,573 \\
Original Rumor Posts & 9,578 \\
ReTruthed Rumor Posts & 89,137 \\
\bottomrule
\end{tabular}
\end{table*}

\begin{figure*}
\centering
\includegraphics[width=\textwidth]{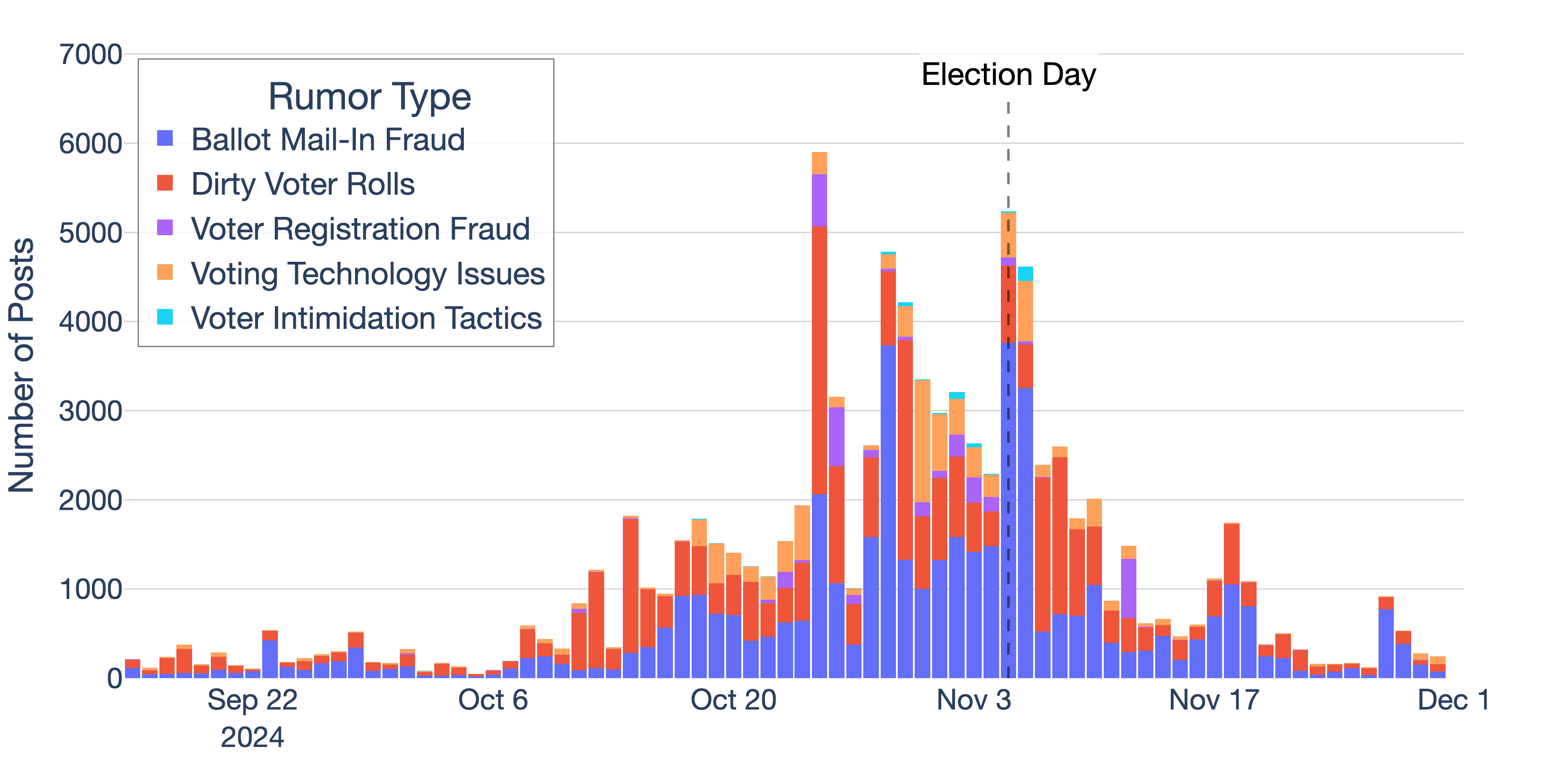}
\caption{Temporal breakdown of rumor posts by rumor type during the key election window.}
\label{fig:temporal_misinfo}
\end{figure*}

Figure~\ref{fig:temporal_misinfo} charts the daily count of rumor posts from 15 September to 1 December 2024, stacked by the five CISA rumor categories. The total column height represents overall rumor volume, while band thickness shows each category's share. Ballot Mail-In Fraud (blue) and Dirty Voter Rolls (red) dominate. A pronounced surge appears from 25 October to 7 November (a period straddling Election Day), with daily election rumor posts climbing to between 4,000 and 6,000 posts, signaling intensified rumor activity as political engagement on the platform peaks, particularly among users aligned with presidential candidate Trump.

To analyze the impact of a user on rumor spreading throughout Truth Social, we define the Influence Metric for a user $u$ as

\begin{equation} \label{eq:influence}
I_u = \sum_{p\in P_u} \Bigl( F_u + \sum_{r\in R_p} F_r \Bigr)
\end{equation}

where $P_u$ is the set of original rumor posts by $u$, $R_p$ is the set of reposts for post $p$, $F_u$ is $u$'s follower count, and $F_r$ is the follower count of each user $r$ who reposts $p$. This metric estimates potential audience reach by summing the author's followers and the followers of all accounts that repost the author's posts. Because follower networks on Truth Social overlap, this metric should be interpreted as an upper bound on potential audience rather than an exact count of unique users. In practice, we use this formulation because follower intersections are not observable for all pairs at our scale, and because the same formulation preserves relative differences across accounts, which is what our analyses use (e.g., ranking users, identifying highly influential accounts). This influence metric, $I_u$, is adapted from the rudimentary reach-based formulation of \citet{Veijalainen2015}, who likewise compute the audience of a singular post as their own followers plus the followers of all users who reshared their message.

To assess whether overlapping follower sets meaningfully affect our conclusions, we conducted two robustness checks. First, we progressively down-weighted the audience contributed by reposters using an overlap parameter $\theta \in \{0.25, 0.5, 0.75\}$, where $\theta = 0$ corresponds to no overlap (our base metric) and higher values simulate increasing overlap. Second, we truncated the audience to only the top-$K$ reposters per post ($K \in \{10, 25, 50\}$), which simulates strong overlap in the long tail of smaller reposters. Across all settings, user-level influence rankings remained highly stable. For overlap-adjusted influence, Spearman correlations with the base metric ranged from $\rho = 0.93$ (at $\theta = 0.75$, the most aggressive assumption) to $\rho = 0.97$ (at $\theta = 0.25$), and 76-80\% of the top-50 accounts remained identical to the base ranking even under conservative assumptions. For reposter truncation, correlations were consistently $\rho \approx 0.97$ across all $K$ values. These results indicate that while the raw influence values constitute an upper bound on potential reach, our comparative findings (which accounts are more influential) are robust to reasonable overlap assumptions. The inflation in absolute exposure values is roughly proportional across users, preserving the relative ordering that our analyses rely upon.

\subsection{Exposure-Propagation Dynamics}
\label{sec:testing_exposure_relationship}
To evaluate our hypothesis (Equation~\ref{eq:hypothesis_exposure}), we identified 51,815 users who received at least one rumor impression. For each of these users, we tracked their neighbors from an information network and identified the timeline of the user's exposures. Using this information, we calculate the probability of a user spreading a specific rumor based on how frequently they have been exposed to the rumor.

To empirically test the individual-level sharing probabilities $P_{u,r}^{\text{share}}(k)$ from Equation~\ref{eq:hypothesis_exposure}, we aggregate across users to compute population-level sharing behavior. For every non-negative integer $k$, we measure the probability that a user has \emph{ever} reshared the rumor \emph{by the time they have accumulated $k$ impressions}. Let $\mathcal{U}$ be the set of users in our panel, $E_u$ the total number of rumor exposures received by user $u$, and $T_u$ the number of exposures that elapsed before $u$'s first (respective-rumor) share. The cumulative sharing probability is therefore:

\begin{equation}
\begin{split}
P^{\text{share}}(k) &= \Pr(T_u \le k \,|\, E_u \ge k ) \\
&= \frac{\sum_{u\in\mathcal{U}} \mathbf{1}\!\bigl[E_u \ge k \,\wedge\, T_u \le k\bigr]}{\sum_{u\in\mathcal{U}} \mathbf{1}\!\bigl[E_u \ge k\bigr]}
\end{split}
\label{eq:cumulative_sharing}
\end{equation}

\begin{figure*}
\centering
\includegraphics[width=0.75\textwidth]{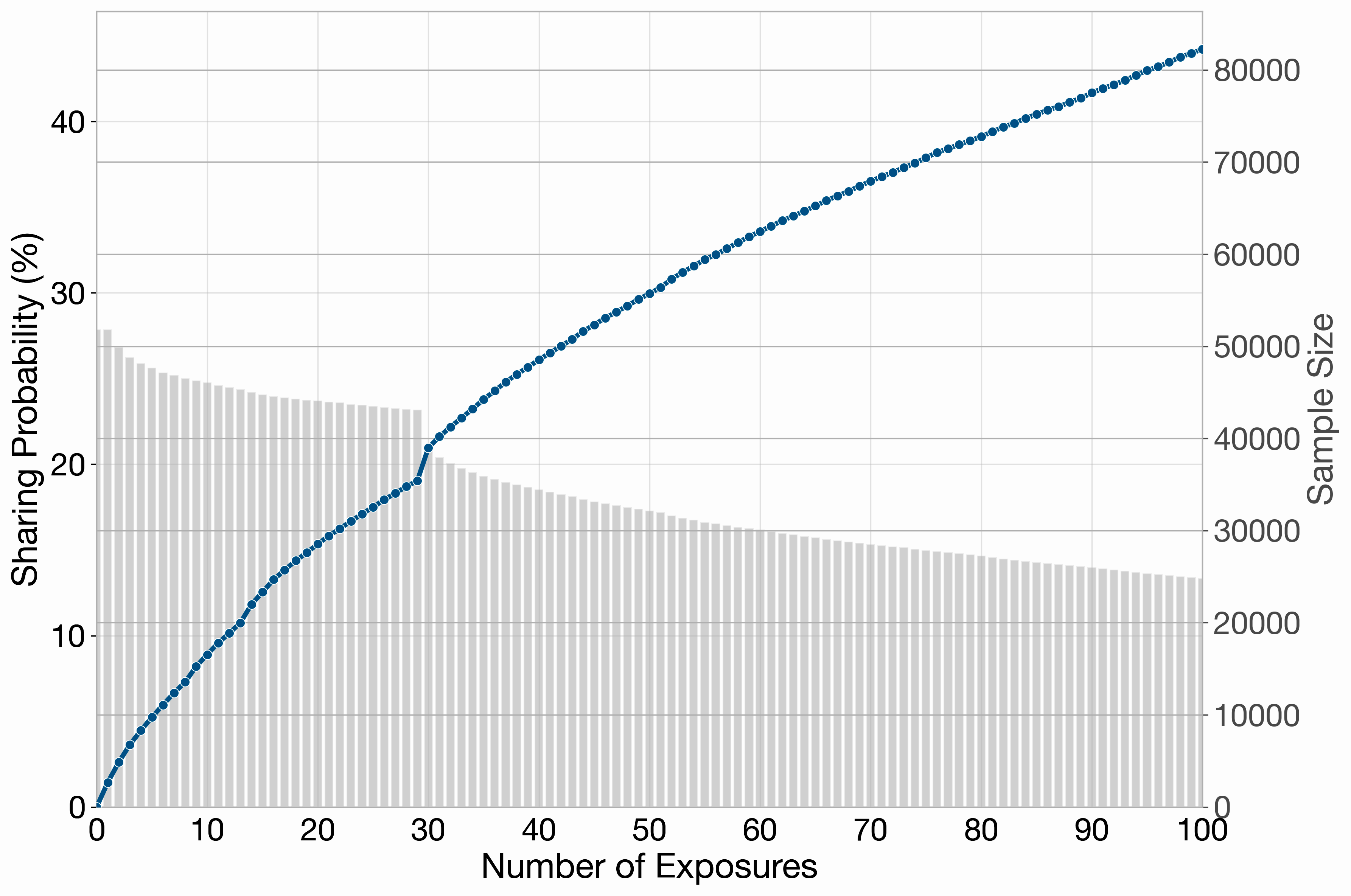}
\caption{Cumulative sharing probability $P^{\text{share}}(k)$ as a function of the number of exposures $k$. Blue line shows sharing probability; grey bars show sample size at each exposure level.}
\label{fig:sharing_probabilities}
\end{figure*}

The denominator represents the population that has \emph{experienced at least $k$ exposures}; the numerator is the subset who have already shared by their $k^{\text{th}}$ exposure. As shown empirically in Figure~\ref{fig:sharing_probabilities}, the curve rises monotonically, indicating a clear dose-response relationship: each additional exposure converts a further fraction of the still-uninfected (``susceptible'') users, with no observed saturation in the range of our data. This finding aligns with psychological research on the illusory truth effect, where repeated exposure to false information increases perceptions of its accuracy \citep{Pennycook2018,Vellani2023}.

Figure~\ref{fig:sharing_probabilities} plots the cumulative sharing probability $P^{\text{share}}(k)$ (Eq.~\ref{eq:cumulative_sharing}). The blue line traces the fraction of users who have reshared an election rumor given that they have had $k$ or more exposures to that specific rumor. In other words, each upward step in the curve represents the additional slice of users who choose to reshare only once their cumulative exposure reaches $k$ impressions. The light-grey bars indicate the sample size of users who have received at least $k$ exposures. Note that the drop-off in sample size from 29 to 30 exposures is due to Donald Trump having exactly 29 rumor posts.

Overall, these findings offer robust empirical support for our hypothesis: on Truth Social, each additional exposure to a rumor substantially increases a user's propensity to share it. This dose-response pattern is emblematic of the illusory truth effect and the tendency for repeated claims to feel more credible \citep{Pennycook2018}. Because Truth Social is an ideologically homogeneous ``echo chamber'', successive exposures are rarely countered by dissenting views; instead they reinforce pre-existing beliefs, lowering skepticism with each repetition \citep{SwireThompson2020HealthMisinformation}. Notably, the curve shows no early plateau; on Truth Social, each new contagion event continues to erode resilience rather than bolster skepticism. In short, users on Truth Social behave less like independent fact-checkers and more like participants in a positive-feedback loop: once a rumor starts circulating, each additional ReTruth further normalizes it, accelerating contagion across the network.

\subsection{Rumor Spread Simulation}
\label{sec:simulation}
To gauge network-level contagion, we construct an information graph $G=(V,E)$ in which each user $u\in V$ is a node and each directed edge $(v\to u)\in E$ represents a repost from $v$ that reaches $u$; edge weights equal the observed repost frequency and thus capture exposure intensity. Every node is assigned a role $c_u\in\{\textit{seed},\textit{spreader},\textit{infected},\textit{ordinary}\}$: \textit{seeds} originate rumors, \textit{spreaders} only repost them, \textit{infected} users have viewed but not yet shared at least one rumor, and \textit{ordinary} users have no exposure.

Following \citet{DeVerna2025}, we simulate diffusion as a weighted threshold cascade. At each discrete step, a node sums incoming weights from seeds and spreaders; once that cumulative exposure exceeds a susceptibility threshold $\phi$, the node flips to the infected state and may transmit the rumor in subsequent rounds. By sweeping $\phi$ from highly susceptible ($\phi=1$) to highly resistant ($\phi=10$) and by allowing heterogeneous thresholds, we trace cascade size and speed, revealing how repeated exposures drive platform-wide spread. This model assumes that exposures accumulate indefinitely without temporal decay or memory effects, which may slightly overestimate the speed and scale of rumor spread. All experiments converged within 4 iterations, highlighting the network's high connectivity. Results from these threshold-cascade simulations are shown in the Supplementary Materials (Figure S3). Across all threshold values, infection remains high, with nearly 14\% of the network adopting the rumor even at the most stringent threshold $\phi = 10$ (requiring 10 direct exposures before becoming infected). Simulations with randomized thresholds produce infection clusters that closely mirror the empirical state-level pattern observed in our data.

\subsection{Geographic Analysis}
\begin{figure*}
\centering
\includegraphics[width=0.75\textwidth]{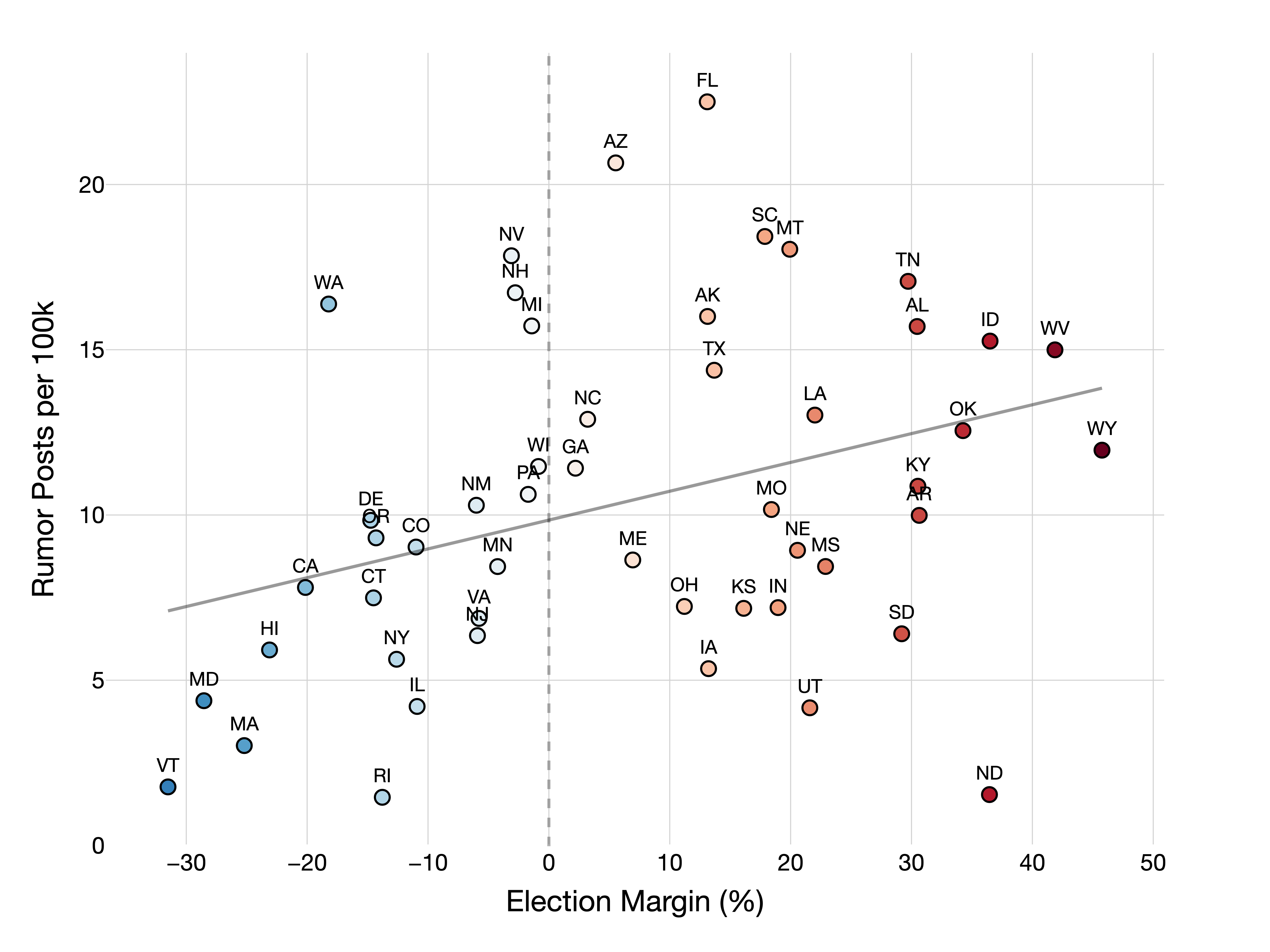}
\caption{Election margin vs. rumor rate by state. Trend line shows positive correlation ($r = 0.34$).}
\label{fig:election_margin}
\end{figure*}

Figure~\ref{fig:election_margin} scatters each state's rumor posts per 100K residents (y-axis) against Trump's 2024 vote margin (x-axis). Point size scales with the number of geolocated users, and a deeper red hue denotes a stronger Trump victory. The positive correlation ($r=0.34$) indicates that users in states giving Trump larger victory margins produce and re-share election rumors at significantly higher rates, consistent with the platform's ideological clustering around his candidacy. However, these geographic patterns should be interpreted with caution, as only 33\% of users were successfully geolocated, and geolocated users may be more active or have richer metadata than non-geolocated users, potentially biasing results toward more engaged segments of the platform. Notable exceptions to the general trend include Florida and Arizona, which show the highest rumor post rates (approximately 23 and 21 per 100K, respectively), and North Dakota, which exhibits one of the lowest rates despite having one of the largest Trump victory margins. The considerable scatter around the trend line indicates that while election margin is a meaningful predictor, other state-level characteristics also contribute to rumor propagation patterns.

\subsection{Central Node of Truth Social: Donald Trump}
Of the 11,000 Trump posts we scraped between May and December 2024, 29 were flagged as rumors; 76\% of those were original rather than reposted content. Two CISA categories dominate (Ballot Mail-In Fraud and Dirty Voter Rolls), together accounting for 86\% of his false claims. Each of his two most influential original posts precipitated measurable platform-wide surges: posts in the same rumor category rose by 35\% in the week following his September 8th claim of ``20\% fraudulent ballots in Pennsylvania,'' and rose by 60\% following his September 17th claim on ``100,000 mislisted Arizona votes.'' Additionally, Trump's cumulative rumor influence score (208.2M) is nearly double that of the next highest user (109.7M), indicating he is responsible for more election rumor spread than any other Truth Social user. The spikes in rumor frequency, coupled with Trump's unrivaled influence in rumor spread, substantiate that he functions as a principal propagator and accelerator of election rumors on Truth Social.

\begin{figure*}
\centering
\includegraphics[width=0.75\textwidth]{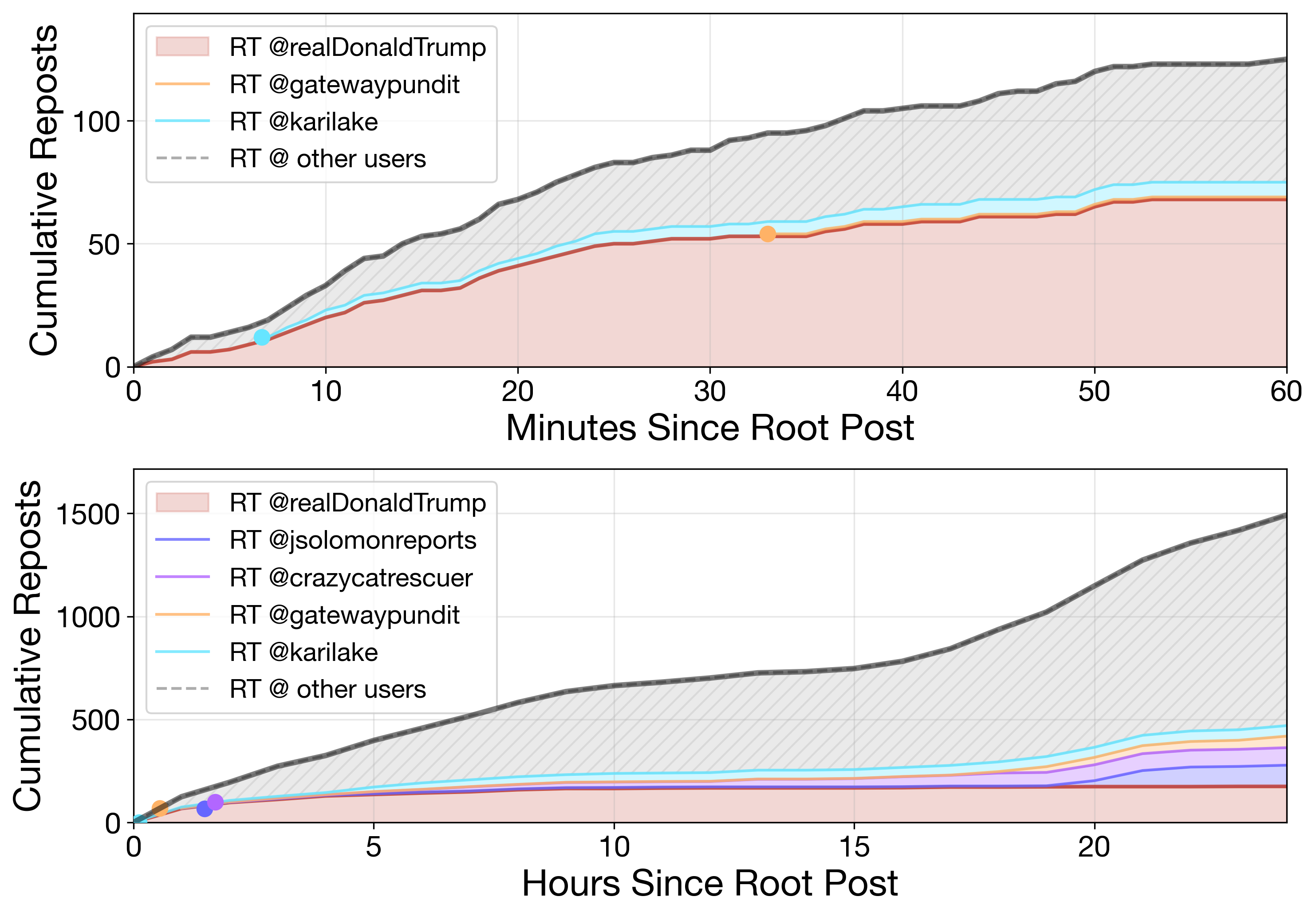}
\caption{Temporal diffusion of Trump's 17 September 2024 Arizona votes claim. The ``root post'' refers to Trump's original post that initiated the rumor cascade. Cumulative ReTruths over the first hour (top) and 24 hours (bottom). Dots indicate when each user first posted their version of the claim. Black line shows total platform-wide ReTruths.}
\label{fig:temporal_diffusion}
\end{figure*}

Figure~\ref{fig:temporal_diffusion} investigates the temporal spread of Trump's September 17th Arizona votes claim by plotting cumulative ReTruths in the 24 hours following his post. In the first few minutes, nearly all repost activity is directly attributable to Trump's original Truth, but within the first hour other high-follower accounts (e.g., Gateway Pundit, Kari Lake) begin posting substantively similar versions of the claim. These secondary posts accumulate their own ReTruths and together push total rumor volume well beyond what Trump's account alone generates. This pattern demonstrates how typical election rumors are spread on Truth Social: a high profile user begins the propagation and it gains traction, before becoming widely posted about by other users (not just reposts) - we can see this as after around 17 hours, many other users begin having posts reposted about this rumor. Rather than remaining concentrated in a single cascade, rumor traffic is re-broadcast by ideologically aligned elites who extend the claim's propagation across the platform.

\section{Discussion}

Our analysis reveals that Truth Social is a potent amplifier for election rumors, with approximately one in 150 posts classified as such. This study, based on a novel dataset of nearly 15 million posts, provides a detailed examination of an alt-tech platform previously under-explored at this scale.

Our work aligns with two major surveys of misinformation detection. First, \citet{Zhou2020} categorize methods into four types: knowledge-based (fact databases), style-based (linguistic cues), propagation-based (network dynamics), and source-based (credibility of origin). Additionally, \citet{Zubiaga2018Survey} further recommend multi-stage rumor pipelines that combine stance detection, veracity classification, and clustering of related content. Looking ahead, recent advances in zero-shot prompting for large language models suggest a promising direction: systems that can identify and adapt to new rumor categories in real-time, with minimal labeled examples.

Key contributions include our multi-stage Rumor Detection Agent — a replicable machine learning and LLM-based framework adaptable for various platforms and rumor types that achieves over 93\% cost reduction through efficient filtering — and our empirical demonstration of a dose-response relationship between rumor exposure and sharing likelihood, consistent with the illusory truth effect. 

Furthermore, we illuminate the structural dynamics of rumor spread, identifying the pivotal role that key political elites can play in accelerating cascades and the propagation of rumors. Our simulation analysis illuminates the speed of likely network saturation, showing that rumors can quickly move through and become widely shared on ideologically homogeneous social media platforms.

Because Truth Social may function as a breeding ground for election rumors, our Rumor Detection Agent could serve as an early-warning trigger for prebunking messages on higher-traffic platforms. Concretely, the agent can surface spikes in a specific CISA category (e.g., ``Dirty Voter Rolls'') within hours; these spikes can trigger platform or agency interventions such as pinning an updated ``Rumor vs. Reality'' item, pushing prebunk content to influential accounts, or limiting resharing of the specific post. Because our detector outputs a structured rumor label, interventions can be chosen per category. In a large UK-based online experiment using a realistic social-media simulation ($n$=2,430), a brief inoculation intervention halved ``likes/loves'' of misinformation (-50.5 \%) and reduced other reactions by 42\% relative to control \citep{McPhedran2023Prebunk}. Linking real-time detection on Truth Social to scalable prebunk delivery would convert the platform from a rumor incubator into a sensor that triggers timely, population-level counter-measures.

However, this study has limitations. Our dataset of 190,000 users covers about one-fifth of Truth Social's estimated one million active monthly users. This is a considerable proportion of Truth Social users. We argue that our sampling methodology should capture and represent nearly all communities on the platform, thus we should have something like a representative sample of Truth Social users during this time period. But as we are not representing the entire universe of Truth Social users during this period, low-incidence or low-activity groups could remain underrepresented. State-level geolocation was achieved for only a third of users, and with state-level granularity, partially limiting the analysis of location-based factors. Additionally, while our Rumor Detection Agent is highly accurate, its initial filtering stages might miss nuanced rumor variants. We did not benchmark against prior LLM-based or propagation-based rumor detectors, nor did we run inter-platform experiments on Gab, Parler, or Telegram. Existing models are typically developed and reported on Twitter-style datasets that provide full conversation trees and richer metadata than our Truth Social crawl, so their published scores would not be informative for our setting. A fair comparison would require reimplementing and redesigning those models under the Truth Social constraints and environment, which we leave for future work. A limitation of our influence estimate is that it does not deduplicate overlapping follower sets. As a result, the absolute exposure values should be read as indicative or magnitude-level rather than literal reach counts. However, because all users are evaluated with the same rule, the metric remains suitable for identifying more versus less influential accounts, as demonstrated by our robustness analysis. The unique nature of Truth Social as a political-focused niche echo chamber also means findings may not directly generalize to broader platforms like X, though they offer insights for similar closed environments. Finally, like many similar studies of social media information dynamics at scale, we have limited information on individual users; future research should continue to find ways to link or develop more information about individual users so as to better understand which users are most affected by rumor spread. 

Currently, we know little about how content moves between social media platforms. For example, do users exposed to rumors on a niche platform like Truth Social take those rumors to other, larger, and more general population social media platforms like Facebook or X? Or does the propagation of these rumors on platforms like Truth Social remain within the confines of its relatively ideologically homogeneous network? There is little research on the role that niche and narrow-cast platforms like Truth Social, Discord, Telegram, or WhatsApp might play in the spread of rumors and falsehoods. How rumors spread across social media platforms is an important question for future research.

\section{Conclusion}

This study demonstrates the transformative potential of Large Language Models (LLMs) for psychological science, specifically in measuring and analyzing the spread of misinformation in large-scale social networks. By developing a multi-stage Rumor Detection Agent that integrates LLM verification with traditional machine learning, we were able to process nearly 15 million posts and provide the first comprehensive quantitative analysis of election rumors on Truth Social.

Our key findings include:
\begin{enumerate}
    \item Election rumors constitute approximately 0.67\% of Truth Social content, with persistent circulation throughout our observation period.
    \item Repeated exposure dramatically increases propensity to reshare, with users showing nearly 10\% increase in resharing probability after 10 or more exposures. This provides large-scale empirical support for the ``illusory truth effect''.
    \item Network simulations demonstrate rapid propagation potential, with 25\% of users potentially exposed within four iterations.
    \item Donald Trump serves as the central node in the platform's rumor ecosystem, driving 40-60\% increases in election rumor discussion.
    \item Geographic clustering reflects local political contexts, with higher engagement in states with contested election results.
\end{enumerate}

These findings have important implications for understanding rumor dynamics on alternative social media platforms and developing effective intervention strategies. The rapid propagation patterns we observe suggest that early detection and intervention are crucial for limiting rumor spread. At this point, we do not know whether niche platforms like Truth Social provide ecosystems where election rumors can be launched and spread, and if rumors then might propagate to more general population social media platforms. What the viral spread of election rumors may be in the larger network of social media platforms is an important area for future research.

Our methodological contributions highlight the value of LLMs for large-scale social analysis. The Rumor Detection Agent provides a reusable, scalable, and cost-efficient framework (reducing LLM calls by over 93\% through filtering) that can be adapted for future research on belief dynamics across different platforms. By enabling the precise measurement of complex social constructs like rumor exposure at scale, such tools open new frontiers for research in political psychology and computational social science.

Future research should extend this analysis to examine cross-platform rumor flows, investigate the effectiveness of different intervention strategies, and develop more sophisticated models of user behavior in rumor ecosystems. Understanding how rumors evolve and spread across different platform environments will be crucial for maintaining information integrity in democratic societies.

\section*{Data availability}
The datasets generated and analysed during the current study are not publicly available due to platform terms of service and user privacy considerations. Post identifiers and metadata sufficient for replication are available from the corresponding author upon reasonable request.

\section*{Code availability}
The Rumor Detection Agent code, sample data, and documentation for replicating the analysis are available at \url{https://github.com/etmaca5/RumorDetectionAgent}.

\section*{Acknowledgements}
The authors thank The Ronald and Maxine Linde Center for Science, Society, and Policy (LCSSP) and the California Institute of Technology for their support.

\section*{Competing interests}
The authors declare no competing interests.

\section*{Ethical approval}
This study was granted exemption from requiring full ethics review by the Institutional Review Board at the California Institute of Technology (IR24-1473). The exemption was granted because the research involves the collection and analysis of publicly available social media data that does not include private or sensitive information, and poses no more than minimal risk to participants.

\section*{Informed consent}
Informed consent was not required for this study. The research exclusively analysed publicly available social media posts from Truth Social, a public platform where users voluntarily share content with the expectation that it may be viewed by others. No private messages or restricted content were accessed. Individual users were not contacted, and no identifying information is reported that would allow individual participants to be identified. The Institutional Review Board at the California Institute of Technology (IR24-1473) confirmed that informed consent was not applicable under these conditions.

\section*{Figure legends}

\textbf{Figure 1.} Webscraping user metadata and posts from Truth Social's website. The database schema is detailed in Figure S1.

\textbf{Figure 2.} Rumor Detection Agent used to classify Truth Social posts. The system uses a fine-tuned RoBERTa model and keyword filtering to identify posts with high probabilities of being a rumor. The posts are then sent to an LLM (GPT-4o mini) where few shot prompting is used first to classify the post as rumor. A second prompt then ensures that the post corresponds to at least one specific rumor.

\textbf{Figure 3.} Temporal breakdown of rumor posts by rumor type during the key election window.

\textbf{Figure 4.} Cumulative sharing probability $P^{\text{share}}(k)$ as a function of the number of exposures $k$. Blue line shows sharing probability; grey bars show sample size at each exposure level.

\textbf{Figure 5.} Election margin vs. rumor rate by state. Trend line shows positive correlation ($r = 0.34$).

\textbf{Figure 6.} Temporal diffusion of Trump's 17 September 2024 Arizona votes claim. The ``root post'' refers to Trump's original post that initiated the rumor cascade. Cumulative ReTruths over the first hour (top) and 24 hours (bottom). Dots indicate when each user first posted their version of the claim. Black line shows total platform-wide ReTruths.

\bibliographystyle{plainnat}
\bibliography{references}

\clearpage
\onecolumn

\setcounter{figure}{0}
\setcounter{table}{0}
\setcounter{algorithm}{0}
\renewcommand{\thefigure}{S\arabic{figure}}
\renewcommand{\thetable}{S\arabic{table}}
\renewcommand{\thealgorithm}{S\arabic{algorithm}}

\begin{center}
\vspace*{1in}
\textbf{\Large Supplementary Materials}\\
\vspace{0.4in}
\end{center}

\vspace{0.5in}

\begin{figure}[!ht]
\centering
\includegraphics[width=0.4\textwidth]{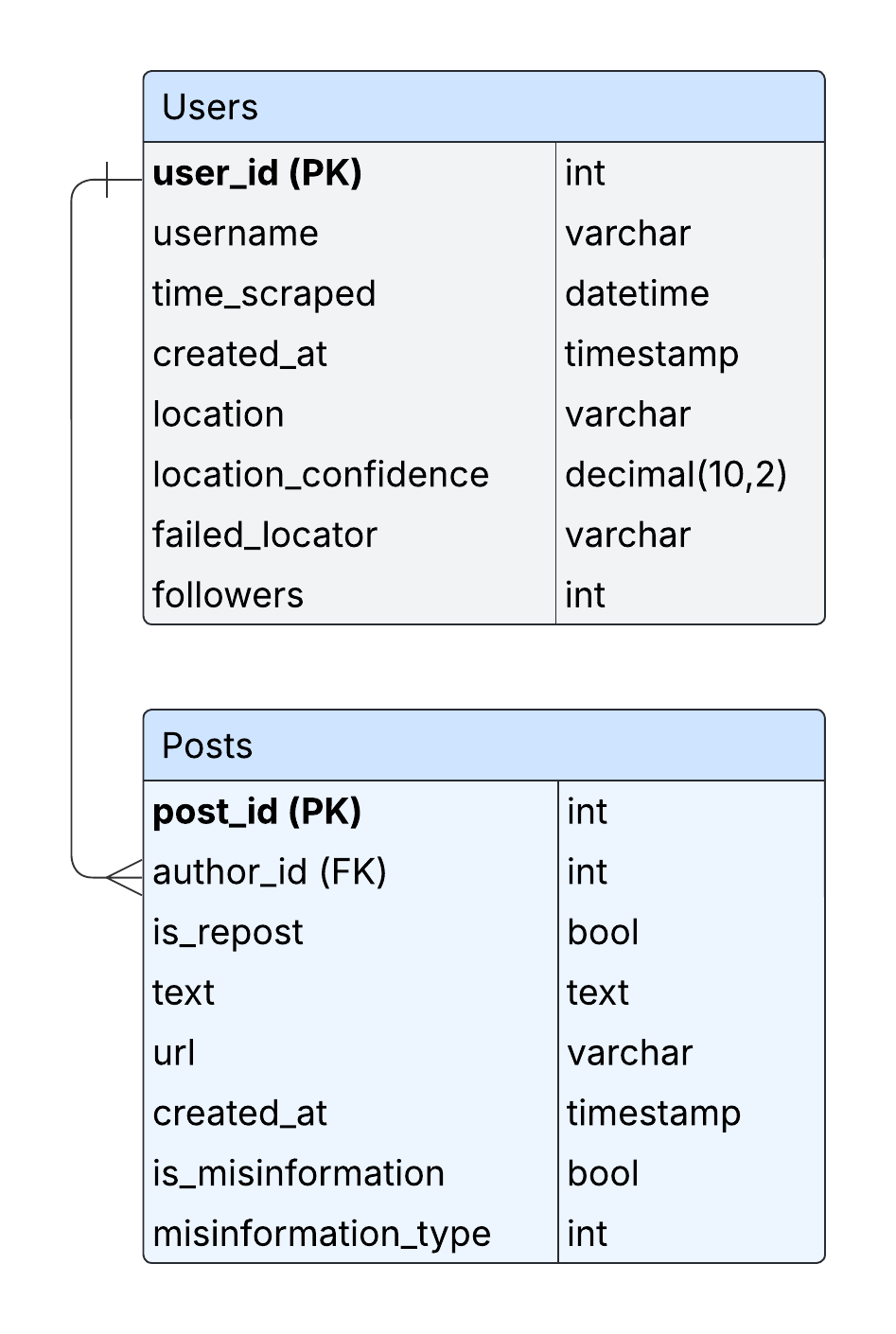}
\caption{Relational Database Schema of Truth Social Election Rumors Dataset.}
\label{fig:database_schema}
\end{figure}

\begin{algorithm}
\caption{User Scraping Algorithm}
\label{alg:user_scraping}
\small
\begin{algorithmic}[1]
\STATE Initialize with a queue containing @realDonaldTrump, and stack for visited users
\WHILE{queue not empty \textbf{and} users less than MAX\_USERS}
    \STATE user $\gets$ queue.pop()
    \STATE Add user to database if not exists
    \STATE Extract follower and following lists
    \FOR{each new follower/following}
        \IF{not in visited set}
            \STATE Add to visited set
            \STATE Add to queue
        \ENDIF
    \ENDFOR
\ENDWHILE
\end{algorithmic}
\end{algorithm}

\begin{algorithm}
\caption{Post Scraping Algorithm}
\label{alg:post_scraping}
\small
\begin{algorithmic}[1]
\STATE Initialize proxies with a pool of IP addresses and accounts. For each proxy:
\WHILE{scraping continues}
    \STATE Query user batch, priority by: (1) Least recently scraped (2) Follower counts
    \FOR{each user in batch}
        \STATE Pull posts created after last scraped time using Truthbrush API
        \FOR{each post}
            \STATE Parse content + metadata, store in database
        \ENDFOR
        \STATE Update user's scraped timestamp
    \ENDFOR
\ENDWHILE
\end{algorithmic}
\end{algorithm}

\begin{figure}[!ht]
\centering
\includegraphics[width=1.0\textwidth]{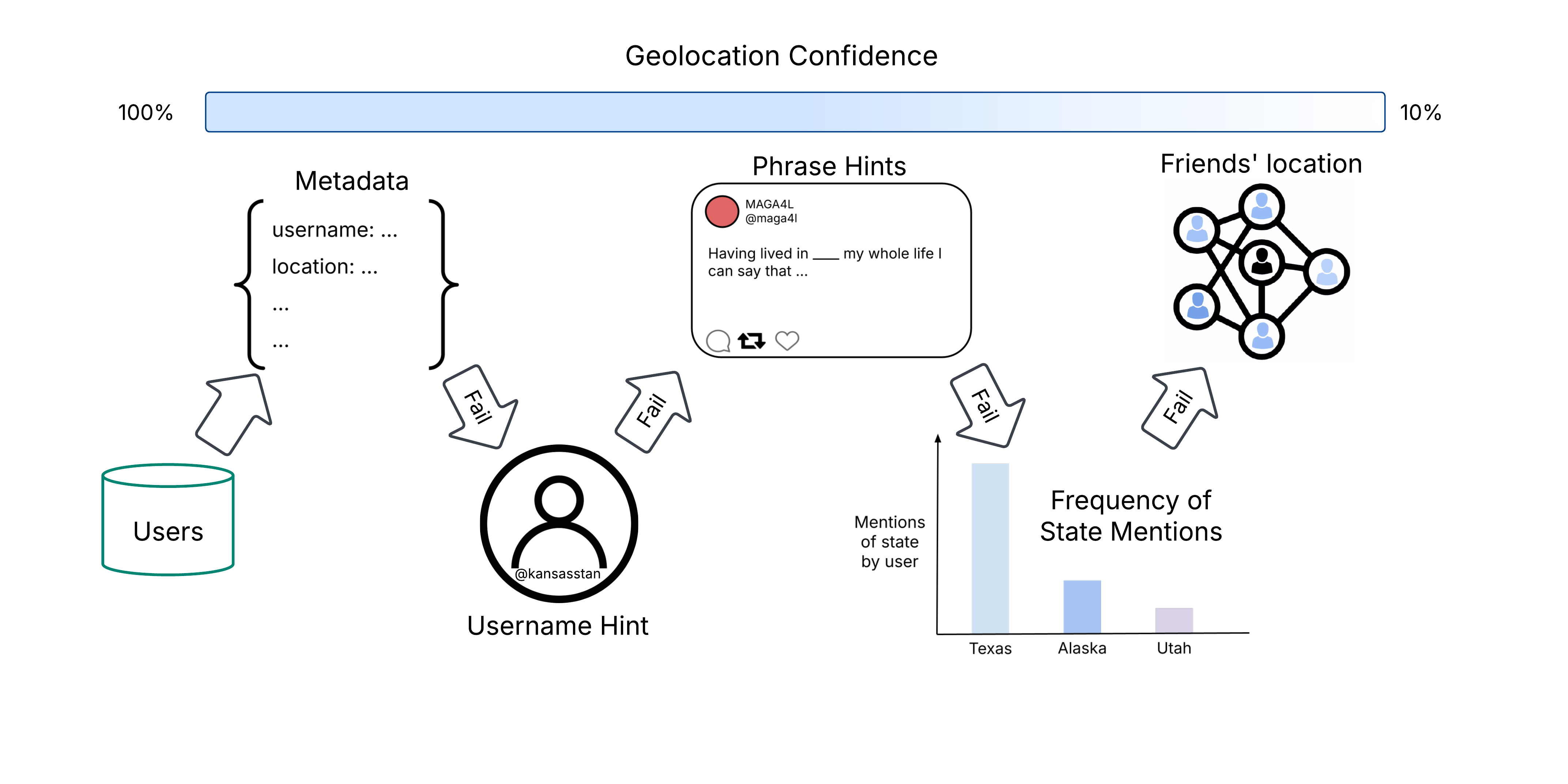}
\caption{Geolocation assignment process for Truth Social users. Attempted assignment descends from high confidence to lower confidence techniques.}
\label{fig:geolocation_process}
\end{figure}

\begin{table}[!ht]
\renewcommand{\arraystretch}{1.2}
\setlength{\tabcolsep}{4pt}
\centering
\small
\begin{tabular}{p{4.0cm}p{7.5cm}}
\toprule
\textbf{Rumor Category} & \textbf{Description} \\
\midrule
Dirty Voter Rolls & Election officials don't clean the voter rolls. The voter rolls are inaccurate and not updated. \\
\midrule
Ballot Mail-In Fraud & People can easily violate the mail-in/absentee ballot request process to receive and cast unauthorized ballots, or prevent authorized in-person voting. \\
\midrule
Drop Box Tampering & Drop boxes used to collect mail-in/absentee ballots can be easily tampered with, stolen, or destroyed. \\
\midrule
Software Security & Voting system software is not reviewed or tested and can be easily manipulated. \\
\midrule
Dead Voters & Votes are being cast on behalf of dead people and these votes are being counted. \\
\bottomrule
\end{tabular}
\caption{Election Rumors}
\label{tab:misinfo_rumors}
\end{table}

\begin{table}[!ht]
\renewcommand{\arraystretch}{1.2}
\setlength{\tabcolsep}{4pt}
\centering
\small
\begin{tabular}{p{4.0cm}c|p{4.0cm}c}
\toprule
\textbf{Binary Classification} & \textbf{Value} & \textbf{Confusion Matrix} & \textbf{Count} \\
\midrule
Accuracy & 94.60\% & True Negatives & 250 \\
Precision & 89.20\% & False Positives & 27 \\
Recall & 100.00\% & False Negatives & 0 \\
F1 Score & 94.29\% & True Positives & 223 \\
False Positive Rate & 10.80\% &  &  \\
False Negative Rate & 0.00\% &  &  \\
\bottomrule
\end{tabular}
\caption{Validation metrics for the Rumor Detection Agent based on an SRS of 500 posts (250 rumor, 250 non-rumor).}
\label{tab:validation_metrics_detailed}
\end{table}

\begin{figure}[!ht]
\centering
\includegraphics[width=0.8\textwidth]{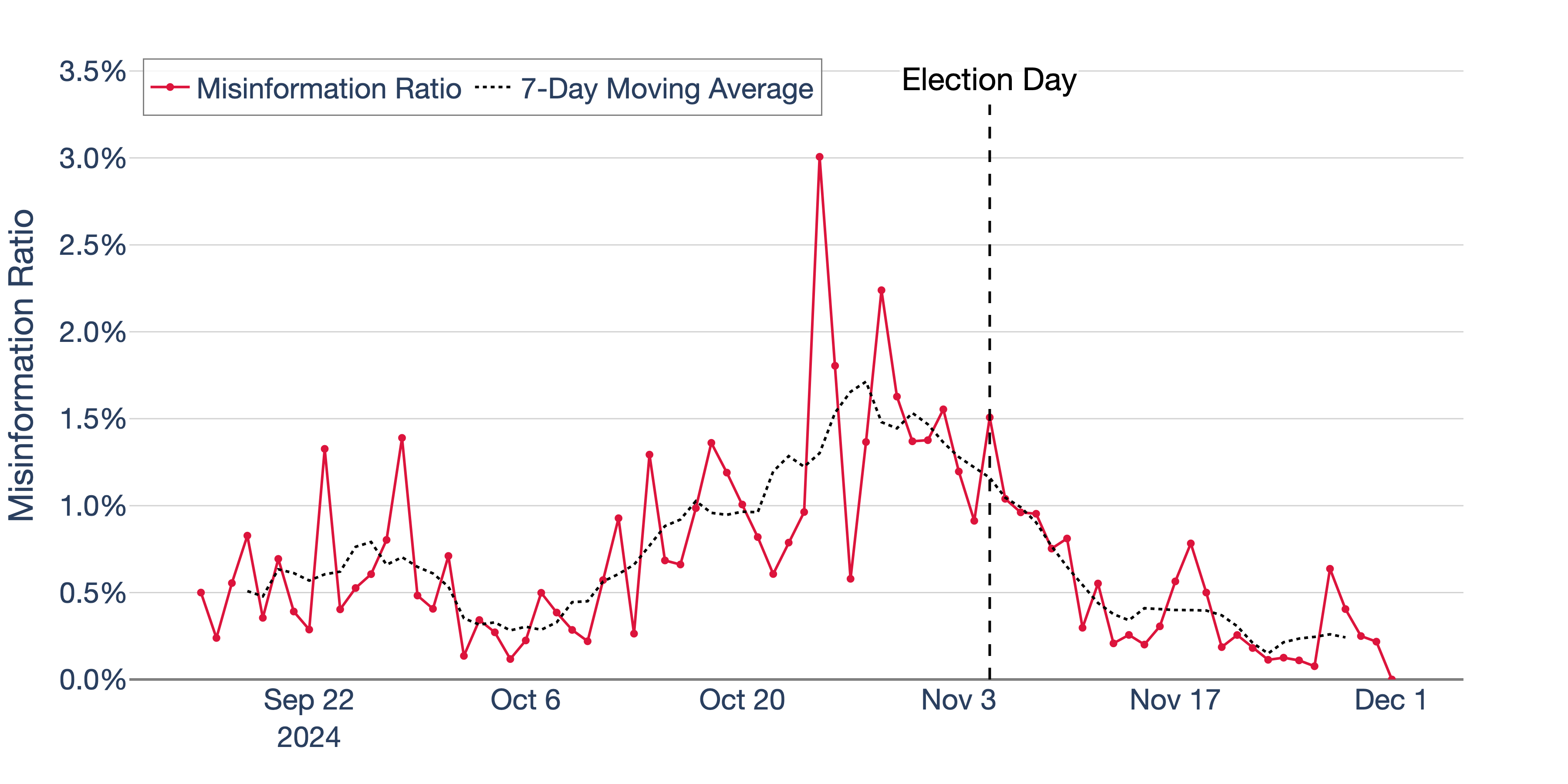}
\caption{Daily frequency of rumor posts on Truth Social during the 2024 U.S. election period. The number of rumor posts peaked in the days leading up to and immediately following Election Day normalized by the total amount of posts in that day.}
\label{fig:daily_freq}
\end{figure}

\begin{figure}[!ht]
\centering
\subfloat[]{%
\includegraphics[width=0.48\textwidth]{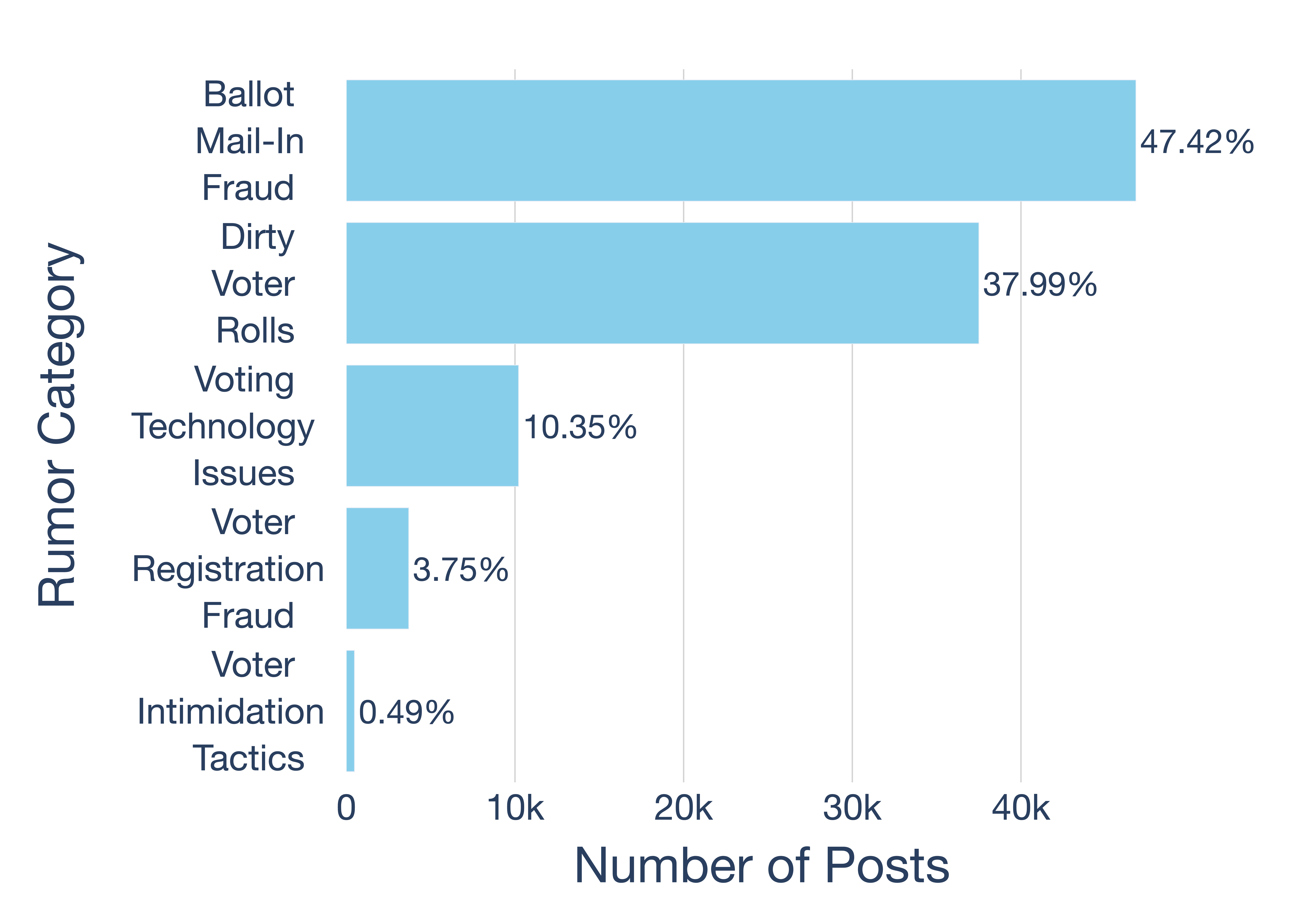}%
\label{fig:misinfo_types}}
\hfill
\subfloat[]{
\includegraphics[width=0.48\textwidth]{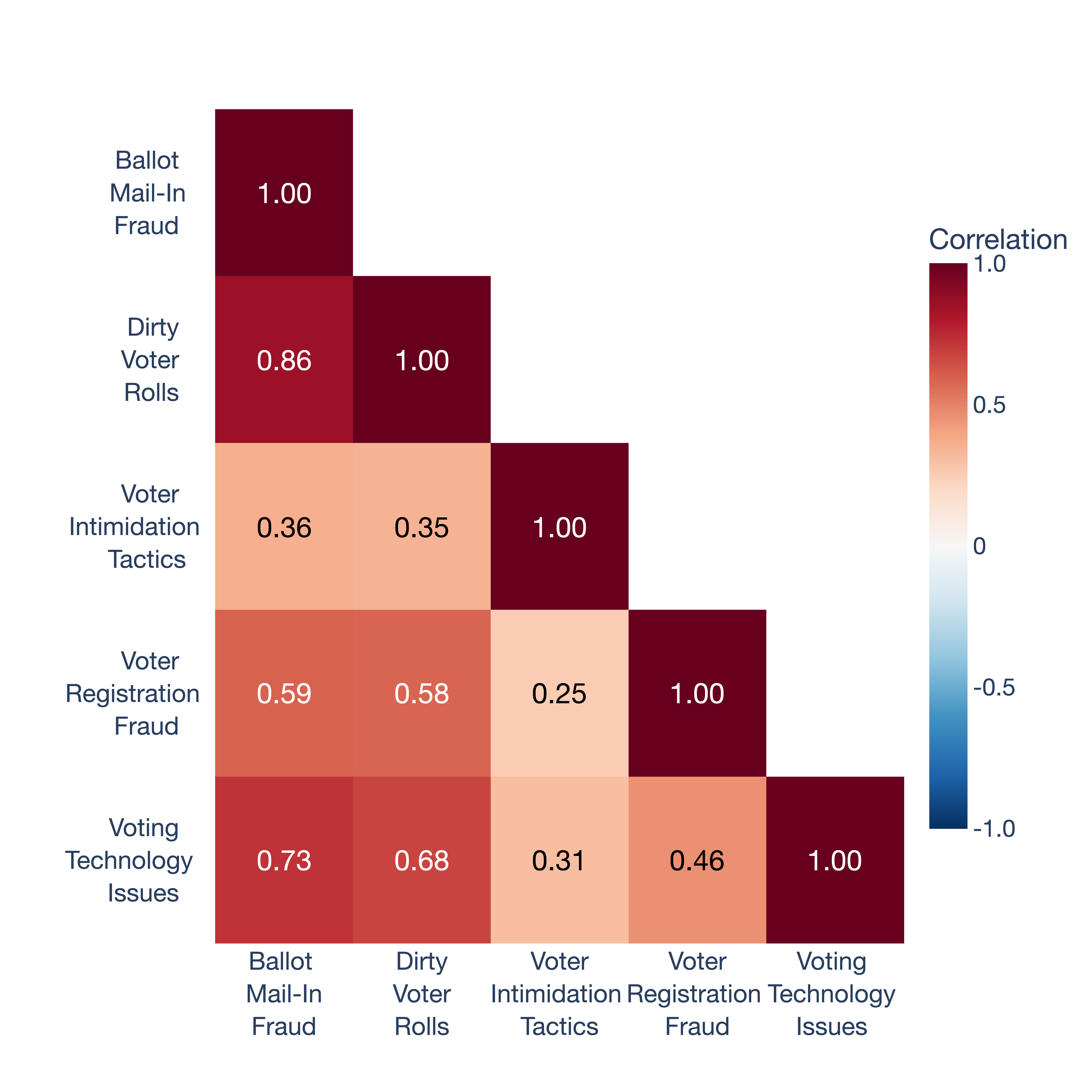}%
\label{fig:misinfo_correlation}}
\caption{Rumor type analysis: (a) Overall frequency of election rumor types. Ballot Mail-In Fraud and Dirty Voter Rolls accounted for the largest share of election rumors, together representing over 85\% of all flagged rumor posts. (b) Correlation matrix between election rumor types. Strong positive correlations indicate users frequently spread multiple types of rumor together.}
\label{fig:misinfo_combined}
\end{figure}

\begin{figure}[!ht]
\centering
\subfloat[]{%
\includegraphics[width=0.48\textwidth]{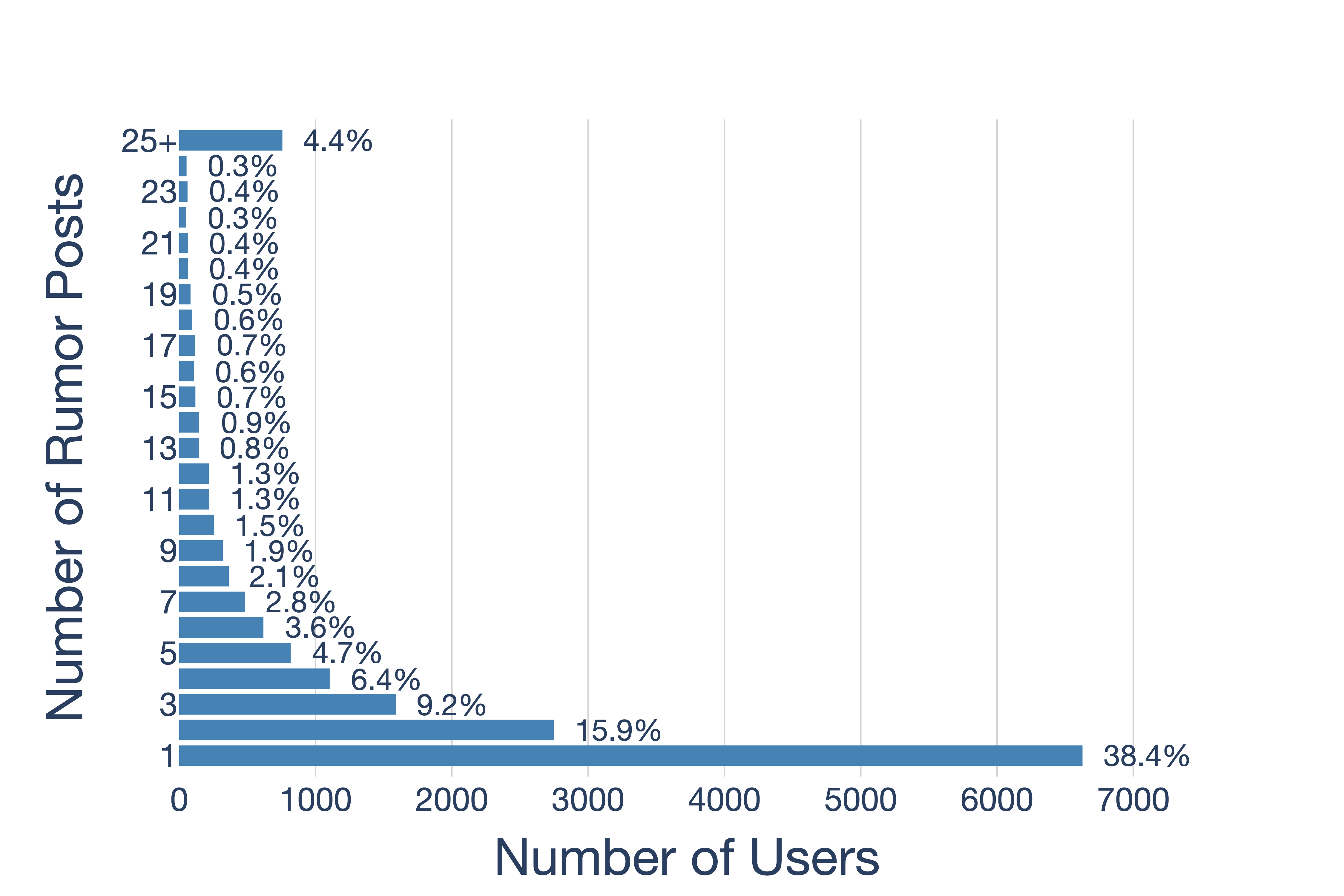}%
\label{fig:user_misinfo_distr}}
\hfill
\subfloat[]{%
\includegraphics[width=0.48\textwidth]{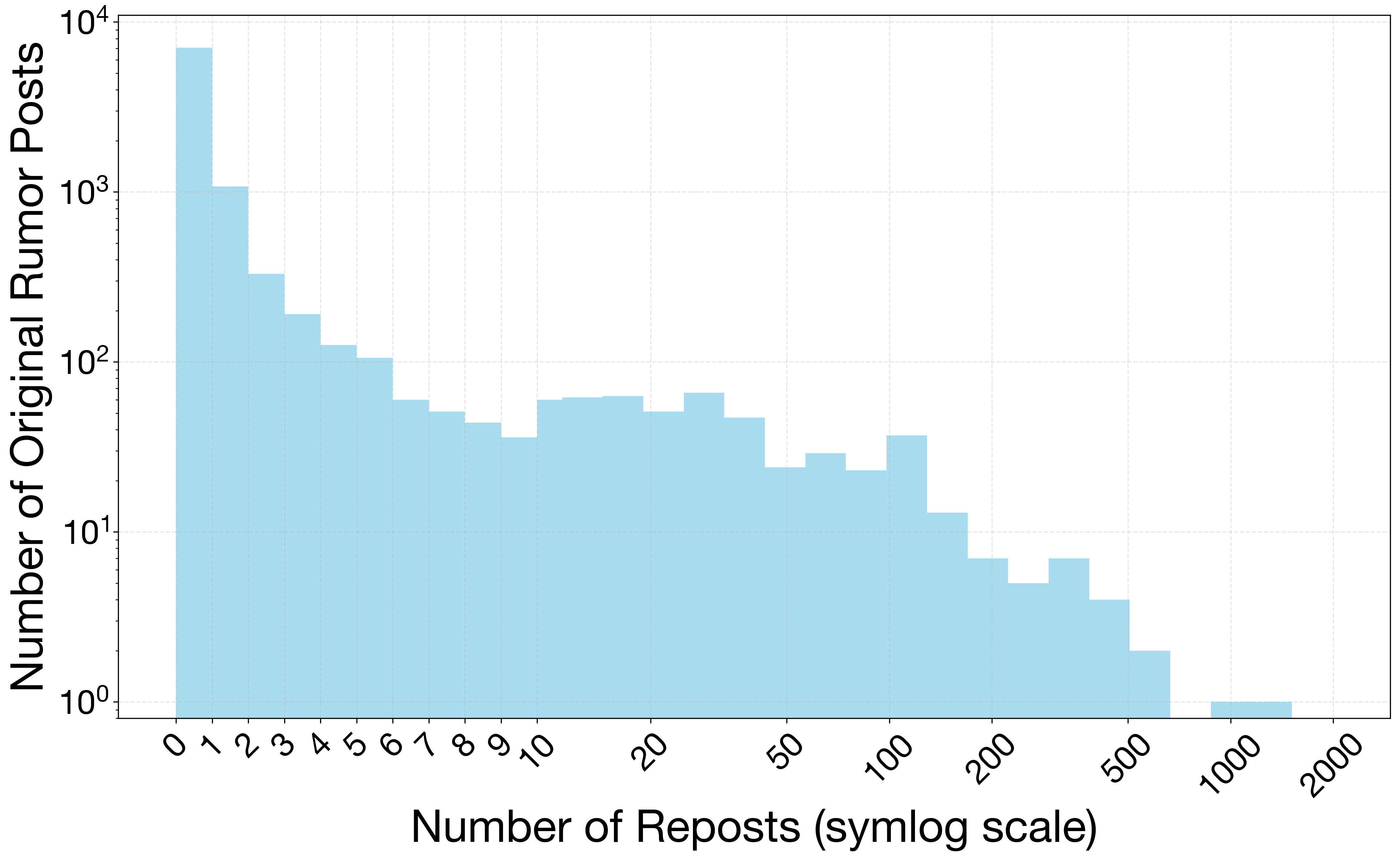}%
\label{fig:distr_misinfo_reposts}}
\caption{User analysis of election rumor: (a) Distribution of rumor posts across users. The majority of users posted election rumors only once or twice, while a small fraction of users (4.4\%) contributed 25+ posts each. (b) Distribution showing the number of reposts on symlog scale (linear near zero, logarithmic for larger values).}
\label{fig:misinfo_user_analysis}
\end{figure}

\begin{figure}[!ht]
\centering
\includegraphics[width=0.9\textwidth]{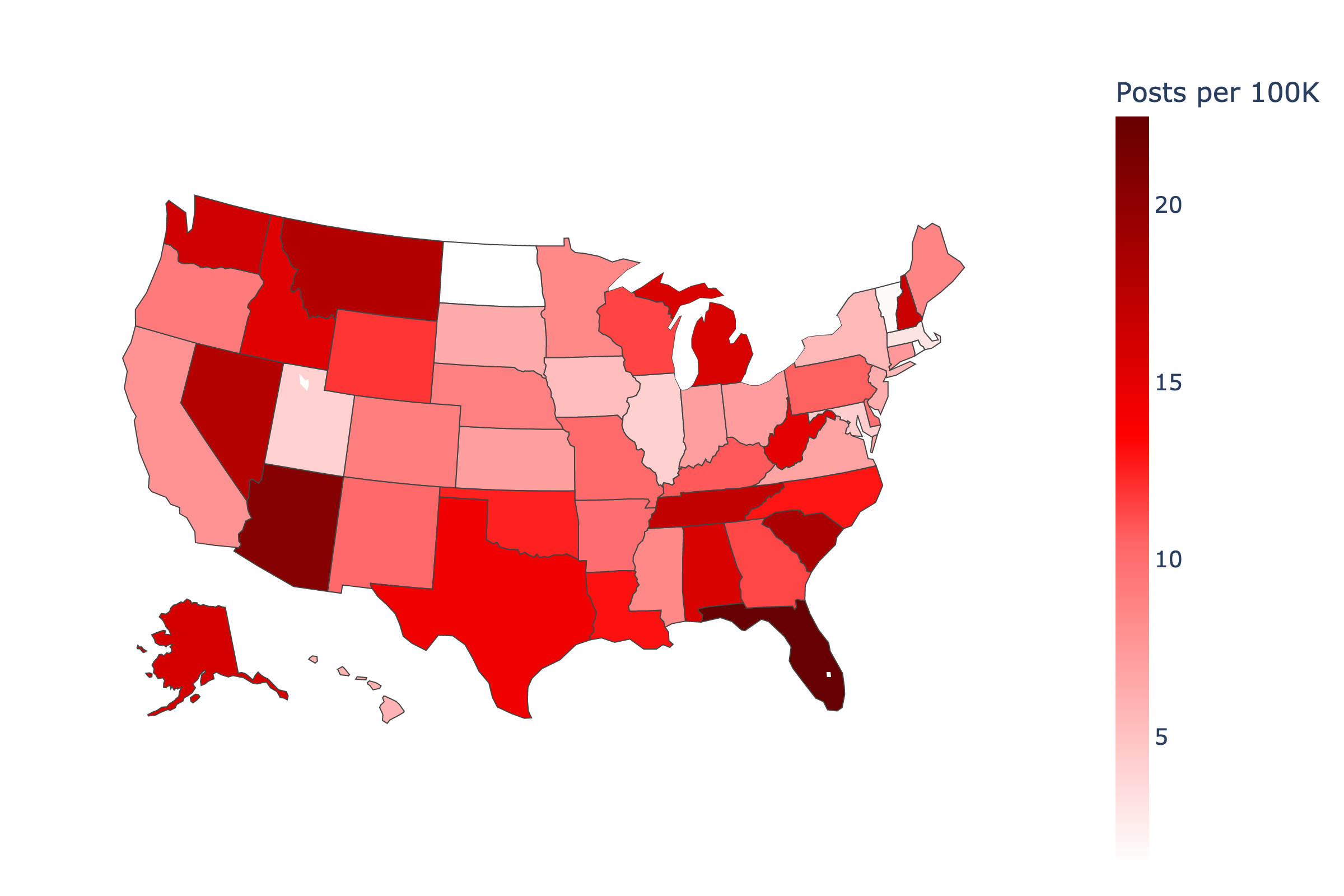}
\caption{Geographic distribution of election rumor posts across the United States, weighted by location confidence scores. The data is normalized based on state population, showing the amount of posts per 100K people.}
\label{fig:misinfo_map}
\end{figure}

\begin{figure}[!ht]
\centering
\includegraphics[width=0.8\textwidth]{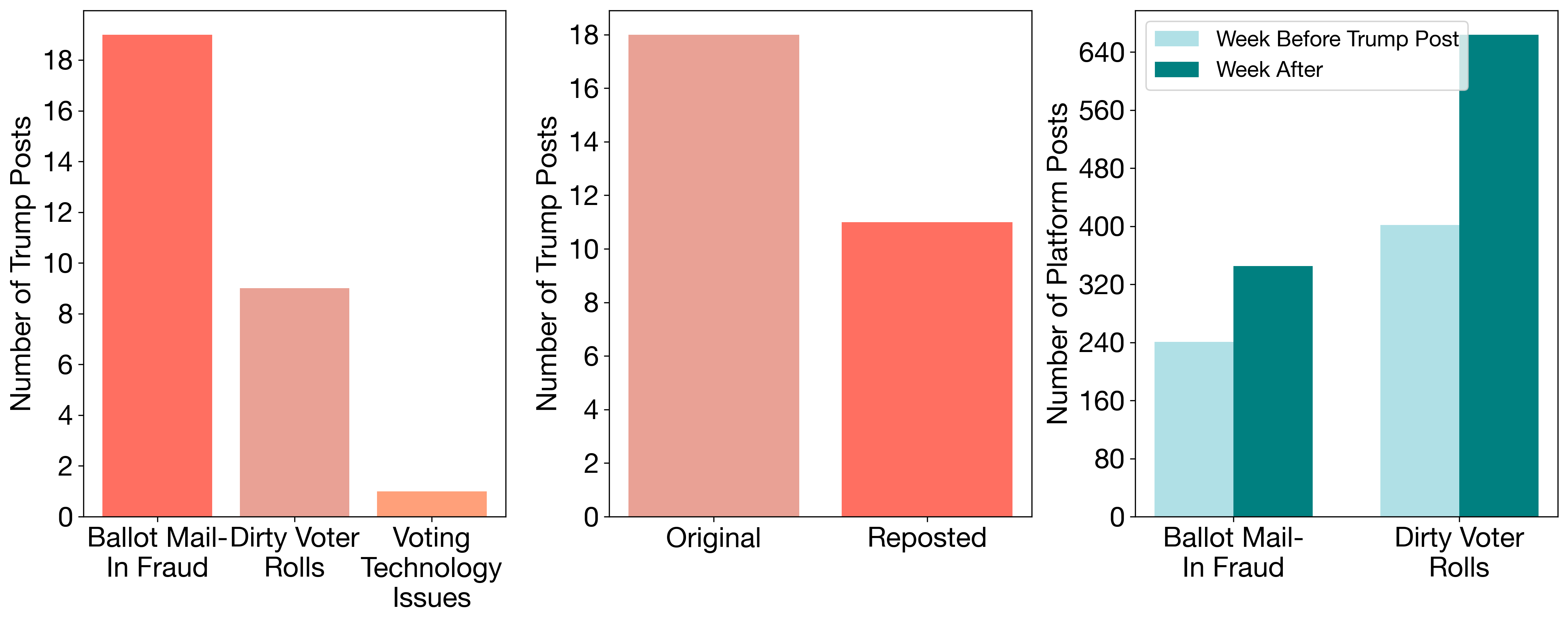}
\caption{Analysis of Donald Trump's rumor spreading activity. (Left) Distribution by rumor category. (Center) Original vs. reposted content breakdown. (Right) Impact of specific Trump posts on platform activity.}
\label{fig:trump_influence_analysis}
\end{figure}

\begin{figure}[!ht]
\centering
\includegraphics[width=0.8\textwidth]{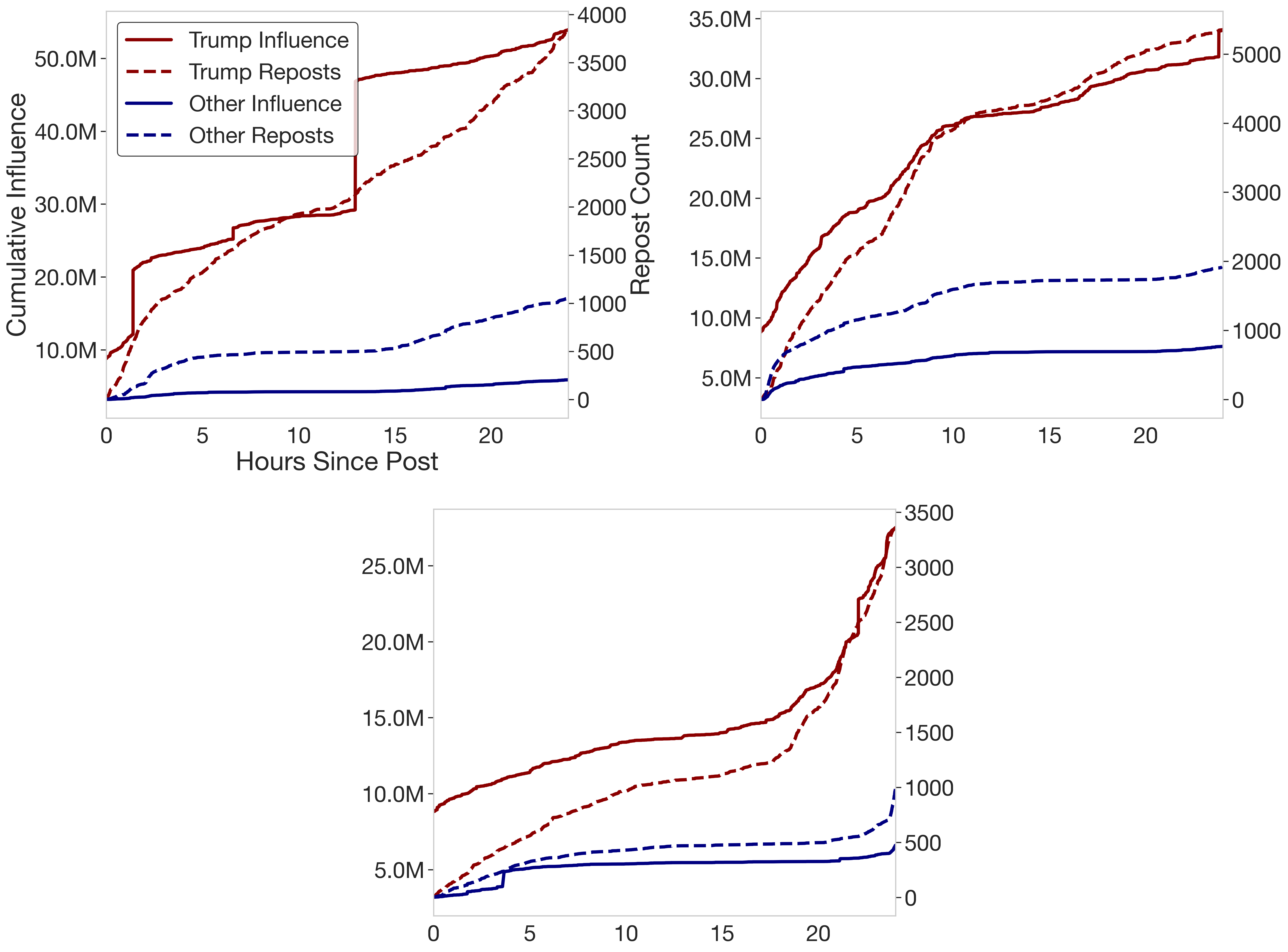}
\caption{Comparison of rumor influence growth: Trump vs. other users. Trump's posts generally start with higher initial influence and grow dramatically faster than other top posts.}
\label{fig:trump_influence_comparison}
\end{figure}

\begin{figure}[!ht]
\centering
\subfloat[]{\includegraphics[width=0.45\textwidth]{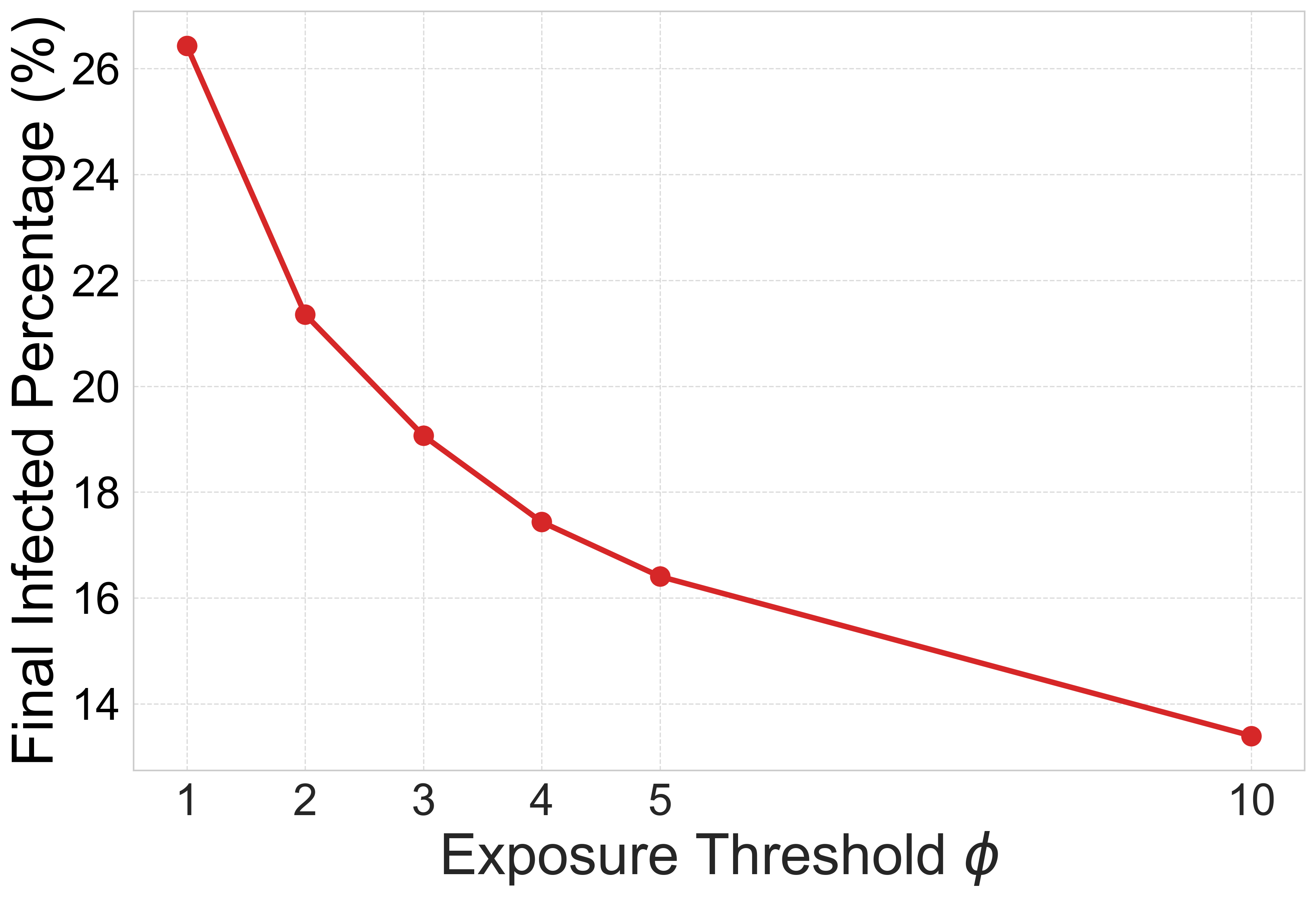}%
\label{fig:misinformed_threshold_vs_misinformed}}
\hfill
\subfloat[]{\includegraphics[width=0.45\textwidth]{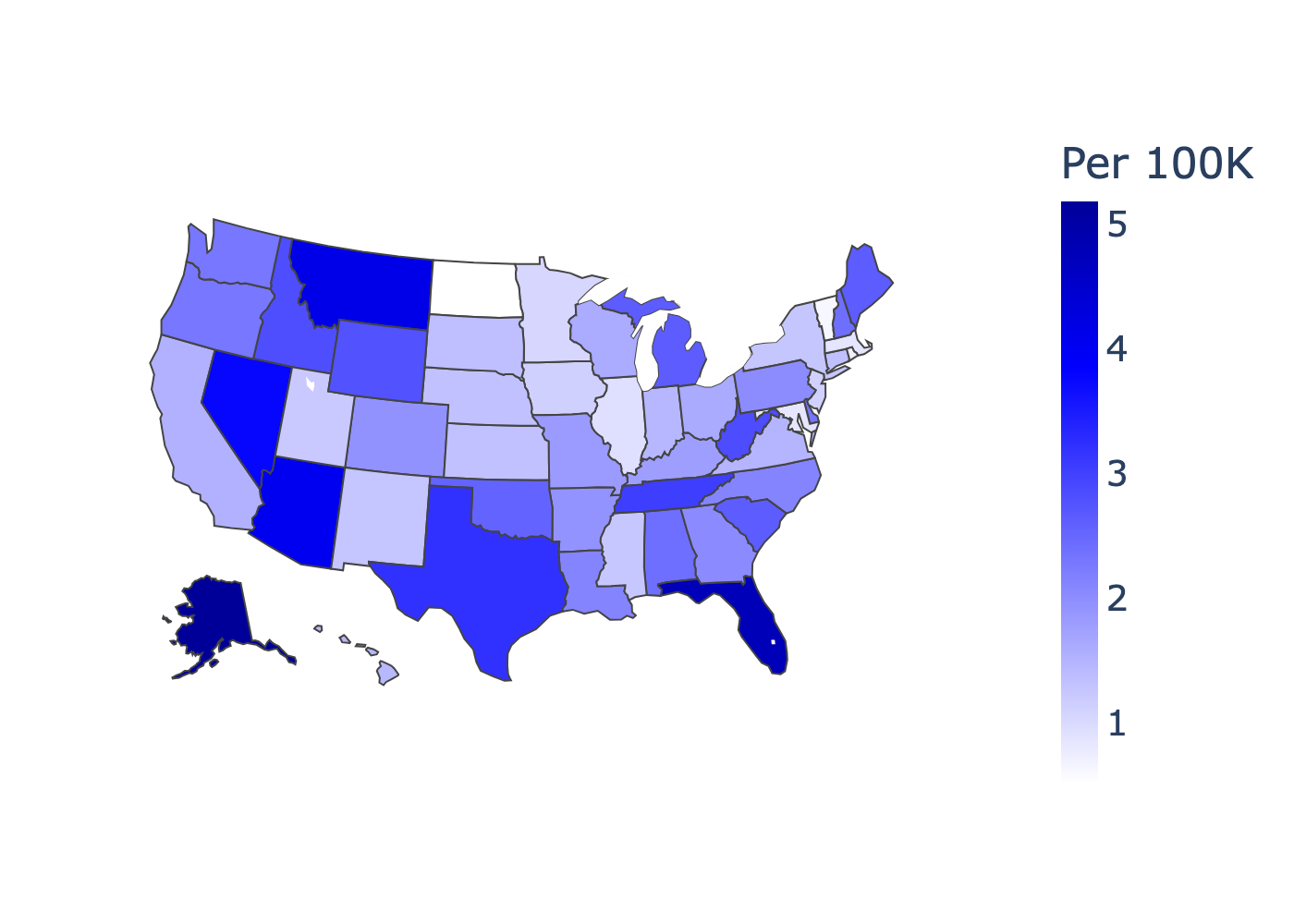}%
\label{fig:misinformed_by_state_map_simulation}}
\caption{Results obtained from the exposure simulations. (a) Exposure threshold $\phi$ vs percentage of infected users at the end of a simulation. Every exposure threshold converged in exactly 4 iterations. (b) Geographic distribution of infected Truth Social Users in an exposure simulation with a randomly sampled exposure threshold for each user.}
\label{fig:exposure_simulation}
\end{figure}

\begin{table}[!ht]
\renewcommand{\arraystretch}{1.2}
\setlength{\tabcolsep}{4pt}
\centering
\small
\begin{tabular}{p{2.5cm}p{2.3cm}p{2.3cm}p{3.5cm}}
\toprule
\textbf{Username} & \textbf{ReTruths} & \textbf{Influence} & \textbf{Description} \\
\midrule
realDonaldTrump & 11,091 & 208,161,832 & President, Owner of Truth Social \\
\midrule
DC\_Draino & 10,385 & 48,111,801 & Conservative attorney \\
\midrule
gatewaypundit & 5,552 & 109,692,478 & Far-right news website \\
\midrule
truethevote & 2,365 & 13,364,985 & Election-integrity organization \\
\midrule
greggphillips & 2,205 & 6,837,993 & Conservative activist \\
\bottomrule
\end{tabular}
\caption{Top users ranked by rumor spread}
\label{tab:top_misinfo_spreaders}
\end{table}

\end{document}